%% file: main.tex
\documentclass[journal]{IEEEtran}

\usepackage{amsmath,amsfonts,amssymb}
\usepackage{algorithmic}
\usepackage{algorithm}
\usepackage{array}
\usepackage[caption=false,font=normalsize,labelfont=sf,textfont=sf]{subfig}
\usepackage{textcomp}
\usepackage{stfloats}
\usepackage{url}
\usepackage{verbatim}
\usepackage{graphicx}
\usepackage{cite}
\usepackage{booktabs}
\usepackage{tabularx}
\usepackage{soul}
\usepackage{makecell}
\usepackage[table]{xcolor}
\usepackage{microtype}
\usepackage[hyphenate]{inconsolata}
\usepackage[hidelinks]{hyperref}

\definecolor{lightgreen}{HTML}{CCFFCC}
\definecolor{lightred}{HTML}{FFCCCC}

\newcommand{\hlg}[1]{{\sethlcolor{lightgreen}\hl{#1}}}

\newcolumntype{L}{>{\raggedright\arraybackslash}X}

\newcommand{\best}[1]{\textbf{#1}}
\newcommand{\dimsc}[1]{\textcolor{gray}{#1}}

\newcommand{\std}[1]{\textsubscript{\textpm#1}}

\providecommand{\citep}[1]{\cite{#1}}
\providecommand{\citet}[1]{\cite{#1}}

\begin{document}
\bstctlcite{IEEEtran:BSTcontrol}

\title{Sentence-Level Attribution for Black-Box Large Language Model\\[-0.5ex] Outputs via Energy-Based Surrogates}

\author{Maryam~Rezaee,
        Pooriya~Safaei,
        Maryam~Asgarinezhad,
        and~Fatemeh~Seyyedsalehi%
\thanks{The authors are with the Computer Science Group at the Department of Mathematical Sciences, Sharif University of Technology, Tehran, Iran\\(e-mail: \href{mailto:ms.maryamrezaee@gmail.com}{ms.maryamrezaee@gmail.com}; \href{mailto:pooriya.safaei@sharif.edu}{pooriya.safaei@sharif.edu}; \href{mailto:maryamasgn123@gmail.com}{maryamasgn123@gmail.com}; \mbox{\href{mailto:seyyedsalehi@sharif.edu}{seyyedsalehi@sharif.edu}}). (Corresponding author: Fatemeh~Seyyedsalehi.)}}

\markboth{Preprint}%
{Rezaee \MakeLowercase{\textit{et al.}}: Sentence-Level Attribution for Black-Box Large Language Model Outputs via Energy-Based Surrogates}

\maketitle

\begin{abstract}
The widespread adoption of proprietary Large Language Models (LLMs) via closed APIs poses an interpretability challenge. While white-box and gray-box (logit-dependent) attribution methods require internal model access, post-hoc attribution under a strictly black-box, text-only generative regime remains largely unexplored. To address this, we propose a model-agnostic, post-hoc attribution interpreter operating at the sentence level. We first train a transformer-based Energy-Based Model (EBM) as a surrogate to score the consistency between prompt and response sentences based on observed LLM generations. To prevent energy collapse and avoid a focus on superficial artifacts, we design contrastive negative samplers that evaluate partial counterfactual dependencies across prompt and response segments. This learned energy landscape then guides the training of a lightweight interpreter network. After model-specific training, the interpreter operates as a standalone tool; it estimates the influence of prompt sentences on a user-specified output sentence without requiring runtime API queries to the target LLM. Comprehensive experiments demonstrate that our EBM surrogate captures nuanced generative dependencies beyond surface-level heuristics, while the interpreter achieves competitive attribution quality with parameter-efficient, $\mathcal{O}(1)$ inference.
\end{abstract}


\section{Introduction}\label{sec:introduction}

\begin{figure*}[!t]
    \centering
    \includegraphics[width=0.75\textwidth]{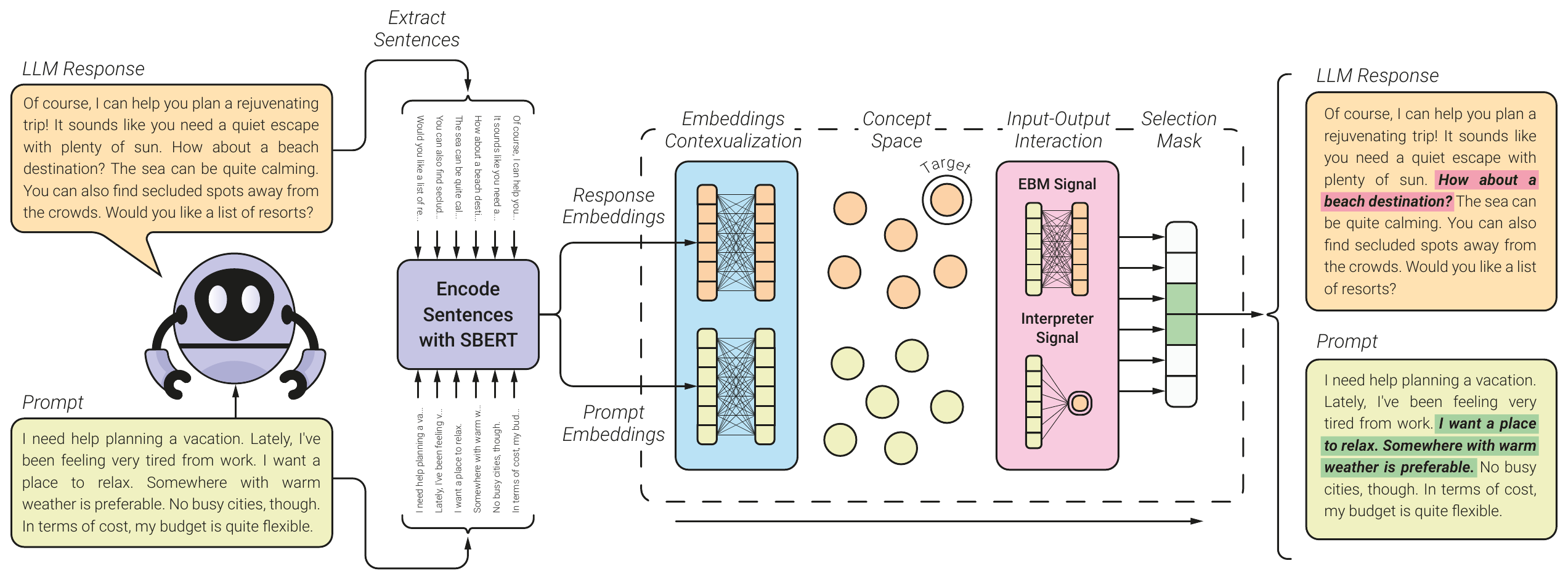}
    \caption{\textbf{Overview of the Proposed Framework.} The prompt and response of a black-box LLM are split into sentences and embedded via a pre-trained model. An encoding module maps these embeddings to a concept space where proximity reflects relevance. An interaction module then evaluates consistency between input and output sentences to identify the most influential prompt sentences for a given target output.}\label{fig:framework_overview}
\end{figure*}

\IEEEPARstart{L}{arge} Language Models (LLMs) demonstrate strong performance across complex tasks and are rapidly being adopted for diverse applications. However, the challenge facing this adoption is the lack of interpretability. Most high-performing LLMs are proprietary and accessible only through closed APIs. Even for open models, scale and complexity obscure how outputs arise. In high-stakes domains like medicine and law, this opacity impedes responsible deployment~\cite{doshivelez2017rigorous}.

Post-hoc attribution is a primary approach to addressing this opacity. These methods explain model behavior by performing instance-wise feature selection---an interpretability paradigm for identifying an importance vector that measures how much each specific input feature influences the prediction for a given instance~\cite{chen2018learning}. However, standard attribution techniques, including white-box and model-agnostic methods, struggle in the context of LLMs. White-box methods, which rely on gradients or activations, are incompatible with closed APIs. Furthermore, the faithfulness of popular proxies like attention weights has been challenged~\cite{jain2019attention}. Model-agnostic methods exist~\cite{ribeiro2016lime, lundberg2017unified, seyyedsalehi2024sointer}, but typically target discriminative models with well-defined outputs. Interpreting generative models is significantly harder as the problem is fundamentally ill-posed. These models utilize complex representations to produce high-dimensional outputs like text. Therefore, effective explanation is hindered by the output's interactivity and sheer volume~\cite{schneider2024explainable}.

Alternatives like prompt-based explanation~\cite{wei2022chain} are similarly problematic; they rely on the same process we seek to verify, leading to circular logic and motivated reasoning. Consequently, models produce plausible yet unfaithful confabulations~\cite{turpin2023language}. While automated prompt engineering can steer model behavior to mitigate biases, it is unsuitable for interpretation. These methods optimize instructions for pre-defined targets~\cite{zhou2023large, clemmer2024precisedebias}, rendering them unusable for ambiguous interpretation tasks.

To address these limitations, we propose a model-agnostic, post-hoc attribution interpreter. We shift the resolution from noisy tokens to sentences, which we define as ``concepts,'' with the goal of relating elements of the output to the prompt at this concept level. While recent mechanistic work defines concepts as latent vectors~\cite{gao2025scaling} or architectural bottlenecks~\cite{sun2025concept}, we adopt a \textit{propositional} view suitable for black-box analysis. Following the Context Principle~\cite{frege1991foundations}, tokens are semantically ambiguous without propositional structure; a sentence, expressing a complete thought~\cite{kintsch1998comprehension}, serves as an operational concept that captures the context necessary for interpretation. Formally, given a prompt $\mathbf{x}$ and an LLM response $\mathbf{y}$, we target a subset of output sentences $\mathbf{y}_T \subseteq \mathbf{y}$ specified by index set $T$. We name the prompt sentences that contribute to $\mathbf{y}_T$ as its \textit{antecedents}, and their degree of contribution as their \textit{influence}. The interpreter then produces an importance vector over $\mathbf{x}$ to isolate the subset of prompt sentences $\mathbf{x}_S \subset \mathbf{x}$ most influential on $\mathbf{y}_T$.

We adopt the paradigm of globally training a local interpreter~\cite{chen2018learning}. Unlike local interpreters that observe only immediate perturbations---which can create interpretability illusions~\cite{friedman2024interpretability}---we train across a diverse set to capture general patterns.

Fig.~\ref{fig:framework_overview} outlines our approach. We first train a transformer-based Energy-Based Model (EBM) as a surrogate for the black-box LLM. The EBM maps sentences into a latent concept space and scores the compatibility between prompt and response sentences based on observed LLM generations. We choose an EBM because interpretation requires evaluating global compatibility without intractable normalization over all possible texts. This energy landscape then guides the training of a lightweight interpreter network. Given a prompt and a target subset of the output, the interpreter identifies the prompt sentences that exerted the most influence on that target.

In summary, our framework makes three contributions: (1) shifting the unit of attribution from tokens to sentences to enable human-intelligible, concept-level analysis; (2) designing a transformer-based EBM with novel contrastive sampling methods to score semantic and structural consistency between prompt and response sentences; and (3) proposing a post-hoc, model-agnostic interpreter that identifies the key prompt sentences influencing a target subset of the LLM's output without requiring inference-time API calls. While our framework focuses on observational attribution rather than causal modeling, identifying these influential antecedents can provide empirical evidence for potential causal relationships in LLM generation.

\section{Related Work}\label{sec:related_work}

\vspace{0.15cm}
\subsection{Post-hoc Interpretation Methods}\label{subsec:posthoc_interpretation}

Without access to internal circuits for mechanistic interpretability, post-hoc attribution remains the primary tool for explaining black-box behavior by scoring input feature importance for a specific model output. White-box approaches to attribution utilize gradients~\cite{sundararajan2017axiomatic, shrikumar17learning, chefer2021transformer} or attention weights~\cite{li2017understanding, xie2017interpretable, hao2021selfattention} to map output signals back to input tokens. However, both are inaccessible via proprietary APIs, and attention is frequently unfaithful to the generation process~\cite{jain2019attention}.

Perturbation-based methods~\cite{ribeiro2016lime, lundberg2017unified, yin2022interpreting} provide a model-agnostic alternative by measuring sensitivity to input alterations, a strategy Hackmann \textit{et al.}~\cite{hackmann2024wordimportance} apply to identify influential words in prompts. Chen \textit{et al.}~\cite{chen2018learning} frames this as instance-wise feature selection by learning an explainer network to maximize mutual information. However, these methods incur prohibitive computational costs for generative tasks~\cite{enouen2024textgenshap, zhao2024reagent}. Consequently, recent work explores training language models to produce answer decompositions as intermediate steps for attribution~\cite{balasubramanian2025decomposition}, or proposes surrogate frameworks to model intermediate generation steps~\cite{chen2026latentdebate}. Unlike these approaches, which necessitate model fine-tuning or internal access, our method performs attribution within strict black-box constraints.

Finally, prompt-based self-explanation~\cite{wei2022chain} prompts the LLM to generate rationales. This diverges from attribution, as it justifies outputs rather than isolating influential inputs. Furthermore, these rationales lack faithfulness guarantees, often serving as plausible post-hoc confabulations~\cite{turpin2023language}.

\subsection{Energy-Based Models in Natural Language Processing}\label{subsec:ebm_nlp}

Energy-Based Models efficiently model high-dimensional structured outputs like text without requiring a normalized probability distribution~\cite{lecun2006tutorial}. While standard probabilistic models must intractably sum to one over all possible outputs, EBMs circumvent this by learning a scalar compatibility score where low energy indicates high data density.

In Natural Language Processing (NLP), EBMs directly refine generation. Residual EBMs add a corrective term to autoregressive log-probabilities to capture high-level coherence~\cite{deng2020residual, bakhtin2021residual}, while other approaches use EBMs to define target landscapes for student network knowledge distillation~\cite{tu2020engine}. Recently, Xu \textit{et al.}~\cite{xu2025energybased} extended these principles to diffusion models, validating that energy landscapes effectively capture the complex distributions of modern text generation.

Beyond generation, EBMs serve as holistic evaluators and post-hoc rankers. Transformer-based discriminators can act as EBMs to distinguish human from machine text by scoring global structure~\cite{bakhtin2019realfake}. This evaluative capacity extends to post-hoc refinement, where EBMs are utilized to rerank candidate outputs and improve generation quality~\cite{bhattacharyya2021reranking}. Most recently, EBMs have enabled robust language model alignment by explicitly modeling reward distributions---instead of standard scalar points---to capture the uncertainty in human preferences~\cite{lochab2025energybased}.

\subsection{Concept-Based Explanations}\label{subsec:concept_explanations}

Interpretability increasingly relies on concept-based explanations to map decisions to human-intelligible ideas~\cite{kim2018interpretability}. While mechanistic methods dominate white-box settings via Sparse Autoencoders~\cite{gao2025scaling} or Concept Bottleneck layers~\cite{sun2025concept}, they require internal access. We address black-box settings by mapping decisions to propositions rather than latent features. This aligns with rationale evaluation standards that favor propositional evidence over isolated tokens~\cite{deyoung2020eraser}. Relying on this, we define each sentence as our conceptual unit, motivated by its ability as a complete thought for interpretation.

Treating sentences as semantic objects has historical precedence. BERT utilized Next Sentence Prediction to learn logical relationships~\cite{devlin2019bert}, while Sentence-BERT confirmed that fine-tuned representations map sentences with similar meanings to proximate points in a semantic space~\cite{reimers2019sentencebert}. By leveraging sentences as concepts, our approach parallels recent architectural innovations like Large Concept Models, which shift from tokens directly to sentence-level representations~\cite{lcmteam2024lcm}.

\section{Methodology}\label{sec:methodology}

\begin{figure*}[!t]
    \centering
    \subfloat[]{\includegraphics[width=0.25\textwidth]{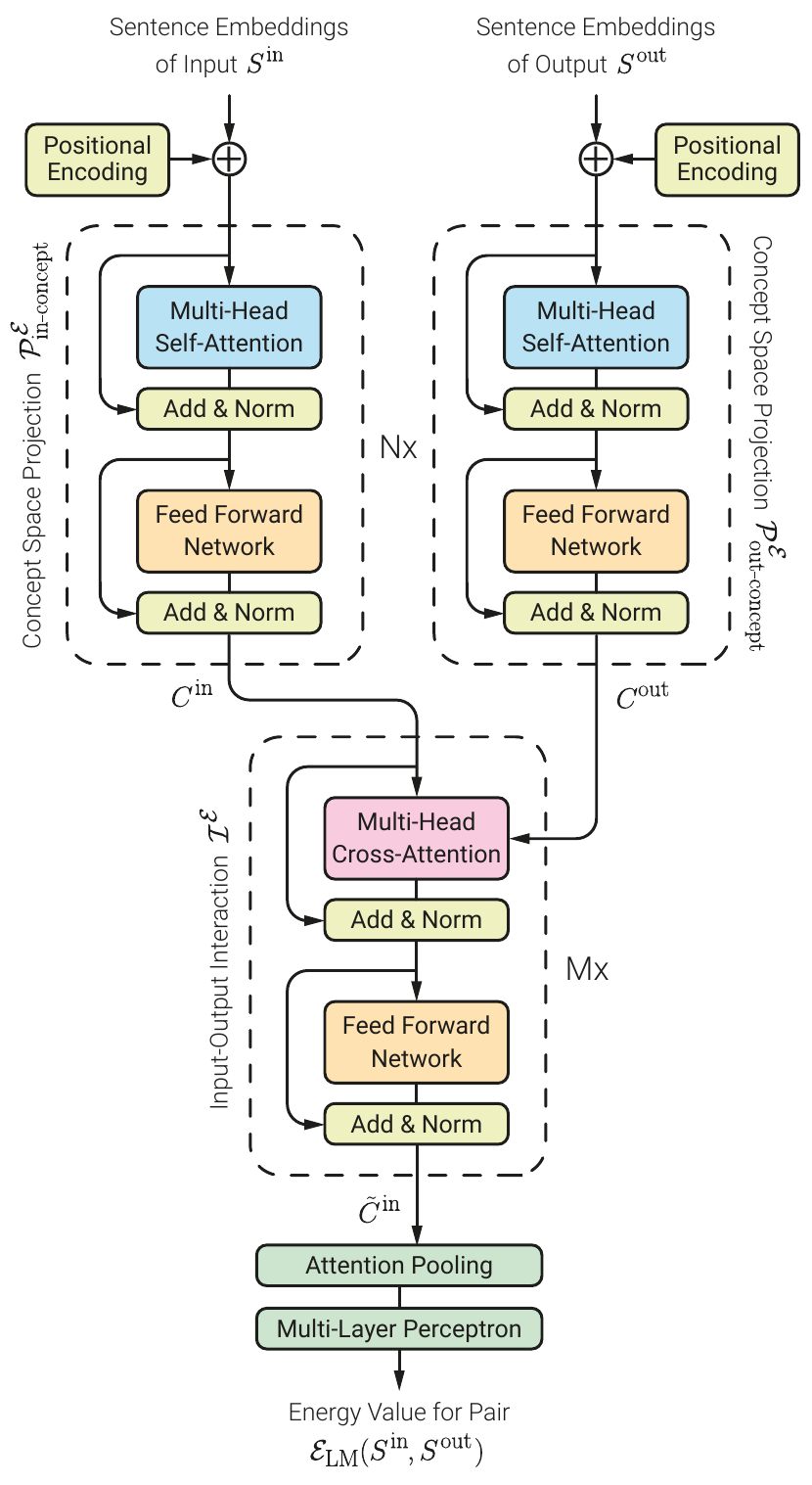}%
    \label{fig:arch_ebm}}
    \hfil
    \subfloat[]{\includegraphics[width=0.36\textwidth]{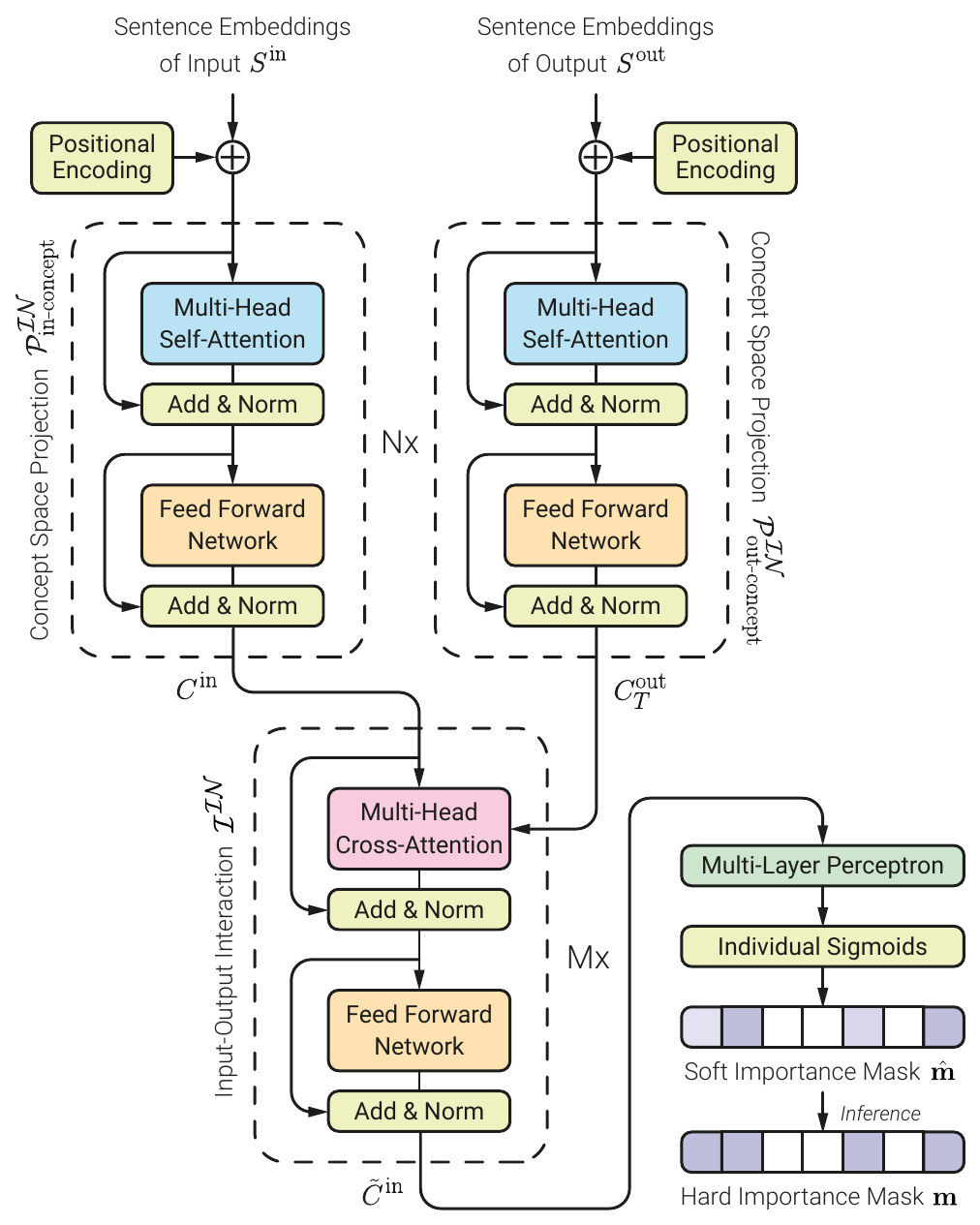}%
    \label{fig:arch_interpreter}}
    \caption{\textbf{Architectural overview.} Schematics for (a) the energy function $\mathcal{E}_{\text{LM}}$ and (b) the interpreter network $\mathcal{IN}$.}\label{fig:architecture_overview}
\end{figure*}

We propose a post-hoc, model-agnostic interpreter for black-box LLMs that identifies the input sentences driving part of a response. We establish the sentence as our analytical unit. As the smallest complete proposition, a sentence serves as an operational concept, framing generation as an interplay of coherent ideas rather than ambiguous tokens. Formally, let $\mathbf{x}$ be the prompt and $\mathbf{y}$ be the LLM response. Given a target subset of output sentences $\mathbf{y}_T \subseteq \mathbf{y}$, where $T$ denotes the set of indices identifying this target subset within $\mathbf{y}$, we estimate the influence of each prompt sentence in $\mathbf{x}$ on the generation of $\mathbf{y}_T$. Our framework operates in two stages. First, we pre-train an EBM, $\mathcal{E}_{\text{LM}}(\mathbf{x}, \mathbf{y}; \theta)$, as a differentiable surrogate for the black-box LLM. As a function of $\theta$, the EBM assigns scalar energy scores representing the compatibility of a prompt-response pair $(\mathbf{x}, \mathbf{y})$ based on the LLM's observed outputs. Second, this energy landscape guides the training of a lightweight interpreter, $\mathcal{IN}(\mathbf{x}, \mathbf{y}, T; \alpha)$, parameterized by $\alpha$. Given the prompt $\mathbf{x}$, the generated response $\mathbf{y}$, and target sentence indices $T$, the interpreter predicts a binary mask over prompt sentences. Values of $1$ identify the subset of prompt sentences $\mathbf{x}_S \subseteq \mathbf{x}$ that serve as key antecedents for the target subset $\mathbf{y}_T$.

\subsection{Sentence Extraction and Embedding}\label{subsec:sentence_extraction_embedding}

The pipeline begins by transforming the input $\mathbf{x}$ and output $\mathbf{y}$ into sequences of sentences. We first perform sentence segmentation using the spaCy library~\cite{honnibal2017spacy}. Subsequently, we employ a frozen, pre-trained Sentence-BERT module~\cite{reimers2019sentencebert} to map each sentence to a fixed-dimensional vector (encoder specifications are detailed in Appendix~\ref{app:ebm_pretraining_details}, Table~\ref{tab:ebm_hyperparameters}). To ensure uniform tensor dimensions ($N_{\text{max}} \times d$) for batch processing, sequences are truncated or padded; notably, we utilize learnable placeholder embeddings to prevent statistical distortion, paired with standard attention padding masks to exclude padded positions from attention computations. This yields embedding sequences $S^{\text{in}}$ and $S^{\text{out}}$, which function analogously to token embeddings within our architecture.

\subsection{The Energy-Based Surrogate Model}\label{subsec:surrogate_ebm}

To approximate the black-box LLM's generation patterns, we design a globally-aware EBM that evaluates entire sequences holistically~\cite{lecun2006tutorial}. Unlike autoregressive token prediction $P(w_t \mid w_{<t})$, the EBM acts as an unnormalized compatibility function. It assigns a scalar score to a prompt-response pair, constructing an energy landscape that reflects semantic and structural consistency between input and output sentences. Because abstract concepts mature hierarchically in deeper layers~\cite{jin2025exploring}, output propositions provide a surrogate for the target LLM's patterns. Thus, the EBM learns to assign lower energy to authentic prompt-response pairs than to corrupted ones, capturing the observed patterns of the target LLM.

As shown in Fig.~\ref{fig:arch_ebm}, the architecture processes sentence embeddings in three stages:

\vspace{0.15cm}
\paragraph{Concept Space Projection} 
Static embeddings ($S^{\text{in}}, S^{\text{out}}$) capture sentence meaning in isolation, lacking full context. We first add positional encodings to provide sequential order. We then pass these augmented embeddings through separate, trainable self-attention modules ($\mathcal{P}_{\text{in}}^{\mathcal{E}}, \mathcal{P}_{\text{out}}^{\mathcal{E}}$) to project them into a contextual concept space ($C^{\text{in}}, C^{\text{out}}$), where proximity reflects relevance within the generation.\vspace{-0.15cm}

\vspace{0.25cm}
\paragraph{Input-Output Interaction}
Because marginal energy models evaluating outputs in isolation fail to capture relational dependencies~\cite{bhattacharyya2021reranking}, we deploy a joint cross-attention block ($\mathcal{I}^{\mathcal{E}}$). Input concepts $C^{\text{in}}$ attend to output concepts $C^{\text{out}}$, yielding response-conditioned representations $\tilde{C}^{\text{in}}$ that verify whether input antecedents are reflected in the response.\vspace{-0.15cm}

\vspace{0.25cm}
\paragraph{Energy Calculation}
The augmented representations are aggregated via attention pooling and passed through a Multi-Layer Perceptron (MLP) to output a scalar energy $\mathcal{E}_{\text{LM}}(\mathbf{x}, \mathbf{y}; \theta)$.

\vspace{0.25cm}
The EBM is trained in two phases: pre-training via a custom sampling strategy, then fine-tuning alongside the interpreter.

For pre-training, we generate a dataset of prompt-response pairs $(\mathbf{x}, \mathbf{y})$ from the target LLM. Because standard contrastive objectives can cause energy collapse in NLP~\cite{deng2020residual}, we optimize a multi-class InfoNCE objective~\cite{oord2018representation} over negative candidates. We optimize two complementary contrastive objectives.

We define the \textit{fidelity} objective ($\mathcal{L}_{\text{fidelity}}$) as an InfoNCE loss to capture overall generation characteristics. By treating responses from humans or other LMs as negative samples, the EBM learns to recognize the target model's output style.

Conversely, we define the \textit{local dependency} objective ($\mathcal{L}_{\text{dep}}$) to target fine-grained semantic connections between prompt and response segments via two batch-wise samplers. The first sampler, $(x_{\text{part}}, y'_{\text{part}})$, targets response dependency for the InfoNCE loss $\mathcal{L}_{\text{resp-dep}}$. Holding a partially masked prompt $x_{\text{part}}$ fixed, it contrasts the true partial response $y_{\text{part}}$ against mismatched partial responses $y'_{\text{part}}$ from the batch. This encourages the model to verify that the output corresponds to that specific input. The second sampler, $(x'_{\text{part}}, y_{\text{part}})$, handles prompt dependency for $\mathcal{L}_{\text{pmt-dep}}$. Holding $y_{\text{part}}$ fixed, it contrasts the true partial prompt $x_{\text{part}}$ against mismatched partial prompts $x'_{\text{part}}$, ensuring the model links the output back to its antecedents.

This piece-by-piece evaluation is critical. Where high-dimensional entanglement obscures dependencies, contrasting counterfactual partial sequences cancels latent noise~\cite{park2023thelinear} while exposing the model to out-of-distribution compositions for robustness. Thus, the objective prevents overfitting to surface heuristics---such as global topic matching or conversational fillers---and forces the surrogate to learn the sentence-level antecedents governing the generation.

Thus, we minimize the combined adaptive loss:
\begin{equation}\label{eq:ebm_total_loss}
    \mathcal{L}_{\text{total}} = (1-\lambda) \mathcal{L}_{\text{fidelity}} + \lambda \mathcal{L}_{\text{dep}} 
\end{equation}
where $\lambda$ is a chosen weight and $\mathcal{L}_{\text{dep}}$ is the sum of $\mathcal{L}_{\text{resp-dep}}$ and $\mathcal{L}_{\text{pmt-dep}}$. We define $h(\mathbf{u}, \mathbf{v}) = \exp(-\mathcal{E}_{\text{LM}}(\mathbf{u}, \mathbf{v};\theta) / \tau)$. Thus, the InfoNCE loss for pair $i$ with negative set $\mathcal{N}_i$ is:
\begin{equation}\label{eq:ebm_infonce_loss}
    \mathcal{L}_i = -\log \left( \frac{h(\mathbf{x}_i, \mathbf{y}_i)}{h(\mathbf{x}_i, \mathbf{y}_i) + \sum_{(\mathbf{x}', \mathbf{y}') \in \mathcal{N}_i} h(\mathbf{x}', \mathbf{y}')} \right)
\end{equation}

Because the quality of the EBM hinges on these contrasts, we detail the specific sampling protocols and hyperparameters in Appendix~\ref{app:ebm_pretraining_details}. This pre-training phase (Fig.~\ref{fig:ebm_pretraining_pipeline}) yields an energy function capturing prompt-response consistency, providing the supervision required to train the interpreter.

\subsection{The Interpreter Model}\label{subsec:interpreter}

Given prompt $\mathbf{x}$, response $\mathbf{y}$, and target sentence indices $T$ (signifying target subset $\mathbf{y}_T \subseteq \mathbf{y}$), identifying influential prompt sentences by evaluating candidates requires an intractable search. Following amortized inference~\cite{tu2018learning}, we instead train a feed-forward interpreter, $\mathcal{IN}(\mathbf{x}, \mathbf{y}, T; \alpha)$, to predict sentence importance in a forward pass, outputting a binary mask where $1$ denotes an influential antecedent sentence.

Fig.~\ref{fig:arch_interpreter} illustrates the architecture, which mirrors the EBM's three stages with targeted modifications. To accelerate convergence, the interpreter initializes its projection modules with the pre-trained EBM weights. As before, positionally-encoded embeddings $S^{\text{in}}$ and $S^{\text{out}}$ are projected into the concept space. In the interaction phase, however, using indices $T$, we retain only target concepts $C^{\text{out}}_T$, discarding non-target output sentences. Input concepts $C^{\text{in}}$ then attend to these targets via cross-attention. Finally, an MLP yields sentence importance logits $\mathbf{z} \in \mathbb{R}^{N_{\text{max}}}$, which a Sigmoid activation relaxes into a continuous mask $\hat{\mathbf{m}} \in {(0,1)}^{N_{\text{max}}}$. At inference, $\hat{\mathbf{m}}$ is binarized using a small threshold $\epsilon$ as $\mathbf{m} = \mathbb{I}(\hat{\mathbf{m}} > \epsilon)$.

During training, however, the continuous mask weights the prompt into a selected subset $\tilde{\mathbf{x}} = \mathbf{x} \odot \hat{\mathbf{m}}$ and an unselected complement $\mathbf{x}-\tilde{\mathbf{x}}$, where $\odot$ denotes element-wise multiplication. Using the frozen EBM as a critic, the interpreter minimizes the energy of $(\tilde{\mathbf{x}}, \mathbf{y}_T)$ to ensure compatibility while maximizing that of $(\mathbf{x}-\tilde{\mathbf{x}}, \mathbf{y}_T)$ to penalize omitted antecedents. Regularized by an $L_1$ penalty to encourage sparse explanations, the training minimization objective is:
\begin{equation}\label{eq:interpreter_objective}
    \resizebox{0.91\linewidth}{!}{%
    $
    \mathcal{L}_{\mathcal{IN}}(\alpha) = \mathcal{E}_{\text{LM}}(\tilde{\mathbf{x}}, \mathbf{y}_T; \theta) - \mathcal{E}_{\text{LM}}(\mathbf{x} - \tilde{\mathbf{x}}, \mathbf{y}_T; \theta) + \lambda_{\text{sparse}} \|\hat{\mathbf{m}}\|_1
    $
    }
\end{equation}

However, conditioning on masked inputs can induce distribution shifts~\cite{hsia2024goodhart}. To mitigate this, we fine-tune the EBM alongside the interpreter via periodic optimization, summarized in Algorithm~\ref{alg:alternating_optimization} (details in Appendix~\ref{app:interpreter_training_details}, Section~\ref{app:interpreter_alternating_optimization}, and Fig.~\ref{fig:joint_training_pipeline}). By querying the target LLM with masked prompts at fixed epoch intervals, fresh generations realign the EBM to the interpreter's evolving distribution. While this joint training incurs API costs, our experiments show it is optional for standard benchmarks yet beneficial for large-scale deployments.

\begin{algorithm}[!t]
\small
\caption{Interpreter Alternating Optimization}\label{alg:alternating_optimization}
\begin{algorithmic}[1]
\REQUIRE Prompt $\mathbf{x}$, Response $\mathbf{y}$, Target Indices $T$ (identifying $\mathbf{y}_T \subseteq \mathbf{y}$), Initialized Interpreter $\mathcal{IN}_\alpha$, Pre-trained EBM $\mathcal{E}_\theta$, Grounding Interval $N_{\text{ground}}$, Threshold $\epsilon$
\vspace{0.15cm}
\FOR{each training step $k$}
    \vspace{0.15cm}
    \STATE \textbf{// Step 1: Interpreter Update}
    \STATE Predict mask: $\hat{\mathbf{m}} \leftarrow \text{Sigmoid}(\mathcal{IN}(\mathbf{x}, \mathbf{y}, T; \alpha^{(k-1)}))$
    \STATE Extract subsets: $\tilde{\mathbf{x}} \leftarrow \mathbf{x} \odot \hat{\mathbf{m}}$
    \STATE Update $\alpha^{(k)}$ via gradient descent to minimize $\mathcal{L}_{\mathcal{IN}}$ (\ref{eq:interpreter_objective}) using frozen critic $\mathcal{E}_{\theta^{(k-1)}}$
    \vspace{0.15cm}
    \STATE \textbf{// Step 2: Periodic EBM Fine-Tuning}
    \IF{$k \pmod{N_{\text{ground}}} == 0$}
        \STATE Apply hard mask: $\mathbf{x}_{\text{hard}} \leftarrow \mathbf{x} \odot \mathbb{I}(\hat{\mathbf{m}} > \epsilon)$
        \STATE Query target LLM: $\tilde{\mathbf{y}} \leftarrow \text{LLM}(\mathbf{x}_{\text{hard}})$
        \STATE Update $\theta^{(k)}$ to minimize (\ref{eq:ebm_total_loss}) for $(\mathbf{x}_{\text{hard}}, \tilde{\mathbf{y}})$, substituting human samples with random base model samples in $\mathcal{N}^{\text{fid}}$
    \ELSE
        \STATE Keep EBM frozen: $\theta^{(k)} \leftarrow \theta^{(k-1)}$
    \ENDIF
    \vspace{0.15cm}
\ENDFOR
\end{algorithmic}
\end{algorithm}

This training process distills the EBM's learned knowledge into the interpreter, thus enabling standalone inference with no reliance on the EBM or need for further LLM API calls.

\section{Experiments}\label{sec:experiments}

Our empirical evaluation investigates two questions: first, does the surrogate EBM ($\mathcal{E}_{\text{LM}}$) capture consistent prompt-response relationships rather than overfitting to surface heuristics? Second, can the downstream interpreter ($\mathcal{IN}$) perform faithful, plausible, and scalable sentence-level attribution under strict black-box constraints? We evaluate across two operational regimes: unconstrained conversational generation (\textbf{HC3}), and constrained factual synthesis (\textbf{WikiBio}).

\vspace{0.15cm}
\paragraph{Target Language Models}
To guarantee that our surrogate approximations do not overfit to one family, we evaluate across three instruction-tuned models with divergent inductive biases, treating each strictly as a closed black box:
(1)~\textbf{\texttt{Llama-3.1-8B-Instruct}}~\cite{grattafiori2024llama3}, a canonical dense model employing Grouped-Query Attention~\cite{ainslie2023gqa};
(2)~\textbf{\texttt{Gemma-2-9B-It}}~\cite{gemmateam2024gemma2}, which incorporates alternating local sliding-window and global attention, GeGLU non-linear activations~\cite{shazeer2020glu}, and logit soft-capping, significantly altering context propagation and predictive entropy; and
(3)~\textbf{\texttt{GPT-4o-Mini}}~\cite{openai2023gpt4}, a proprietary, closed-weights frontier model whose exact architecture and training recipe remain undisclosed, representing production commercial APIs characterized by extensive alignment post-training, conversational preambles, and distinct stylistic priors.

\input{Tables/tab_model_scale_comparison}

\vspace{0.15cm}
\paragraph{Benchmark Datasets and Task Adaptation}
Evaluating post-hoc attribution in open-ended generation is inherently ill-posed because internal computational traces lack deterministic ground truth. We therefore establish two complementary benchmark settings: First, \textbf{HC3}~\cite{guo2023closechatgpthumanexperts} provides multi-domain, unconstrained conversational Q\&A (spanning Reddit ELI5, medicine, and law), where models generate free-form responses to user prompts.
Second, to establish an simpler testbed with clearer antecedents, we adapt \textbf{WikiBio}~\cite{lebret2016neural} by decomposing tabular infoboxes into sequences of isolated, atomic propositions, prompting the target LLM to synthesize a biography strictly from these facts without external knowledge.
Representative samples and prompt formulations for both benchmarks are detailed in Appendix~\ref{app:datasets_adaptation} (Tables~\ref{tab:hc3_dataset_sample} and~\ref{tab:wikibio_dataset_sample}).

\vspace{0.15cm}
\paragraph{Baseline Implementations and Scoping}
We evaluate ESCI against a suite of interpretability baselines, structured into three distinct operational categories:

\vspace{0.1cm}
\noindent{1. Reference Baselines:}
We incorporate two non-parametric reference checks:
(a)~\textbf{\texttt{StringMatching}}: For WikiBio, we construct a quasi-ground truth as a simple heuristic baseline by performing fuzzy string matching between prompt table facts and the generated output sentences (see Appendix~\ref{app:fuzzy_string_matching} for the full algorithmic procedure); and
(b)~\textbf{\texttt{Dirichlet Random}}: Samples sparse importance vectors $\mathbf{s} \sim \text{Dir}(0.05)$ to control for arbitrary sentence selection and mask sparsity.

\vspace{0.1cm}
\noindent{2. Black-Box Methods:}
Operating via prompts and outputs without internal model access:
(a)~\textbf{Oracles}: Five LLMs (\texttt{GPT-4o-Mini}, \texttt{Gemini-2.5-Flash}, \texttt{Llama-3.1-8B-Instruct}, \texttt{Gemma-2-9B-It}, and \texttt{Mistral-7B-Instruct-v0.2}) prompted zero-shot to score prompt sentence importance; a target model's introspection over its own output serves as self-explanation.
(b)~\textbf{\texttt{LIME-BlackBox}}: LIME~\cite{ribeiro2016lime} is a prominent and widely adopted local surrogate framework in explainable AI. Because standard LIME operates on tokens and requires model log-likelihoods (a gray-box setting), we adapt it to sentence-level and formulate a black-box counterpart that scores perturbations via generation degradation (Appendix~\ref{app:lime_sentence_adaptation}).

\vspace{0.1cm}
\noindent{3. White-Box and Gray-Box Methods:}
Although white-box methods operate under a privileged access paradigm fundamentally inaccessible in closed deployments, we benchmark against them on open-weights models to examine how a black-box surrogate compares against internal representations:
(a)~\textbf{\texttt{LIME-GrayBox}}~\cite{ribeiro2016lime}, which fits linear surrogate weights to forced-decoding token log-likelihood drops of the target sentence (Appendix~\ref{app:lime_sentence_adaptation});
(b)~\textbf{\texttt{Raw Attention}}~\cite{vaswani2017attention}, taking mean multi-head attention weights from the final layer to input tokens;
(c)~\textbf{\texttt{Attention Rollout}}~\cite{abnar2020quantifying}, recursively tracking attention flow across successive transformer layers;
(d)~\textbf{\texttt{GradientSHAP}}~\cite{lundberg2017unified, erion2021improving}, computing expected gradients with a single reference baseline;
(e)~\textbf{\texttt{Input} $\times$ \texttt{Gradient}}~\cite{li2016visualizing}, computing the element-wise product of input token embeddings and their gradients with respect to target output likelihoods; and
(f)~\textbf{\texttt{Sentence-LOO}}~\cite{li2017understanding, zeiler2014visualizing}, iteratively masking each prompt sentence and measuring the forced-decoding log-likelihood drop of the target sentence.
To align token-level white-box methods with sentence granularity, token attribution scores are summed across each sentence. For intervention evaluations, we retain sentences with strictly positive attribution scores ($s_i > 0$).

\subsection{Ablation Study: EBM Objectives}\label{subsec:experiments_ebm_ablation}

We ground our bidirectional energy architecture in the findings of Bakhtin \textit{et al.}~\cite{bakhtin2019realfake}, who showed that bidirectional transformers excel at scoring holistic text distributions for EBMs. However, we introduce critical modifications to both the architecture and the contrastive objective (\ref{eq:ebm_total_loss}). Per the design of our dual-objective EBM, a faithful surrogate reconciles two imperatives: \textit{fidelity}, to capture global stylistic consistency of the target LLM, and \textit{local dependency}, to isolate sentence-level antecedents. To justify this formulation, we compare three representative variants: $\mathcal{E}_{\text{Fidelity}}$ ($\lambda=0$, global consistency only), $\mathcal{E}_{\text{Dep}}$ ($\lambda=1$, local dependency without global regularization), and $\mathcal{E}_{\text{Hybrid}}$ ($\lambda=0.9$, balancing both).

\vspace{0.15cm}
\paragraph{Diagnostic Evaluation Framework}
To understand why different EBM objectives succeed or fail as critics, we probe their learned function directly. Using ablation~\cite{li2017understanding}, we evaluate sentence importance across these surrogates by establishing a baseline energy $E_{\text{base}} = \mathcal{E}(\mathbf{x}, \mathbf{y})$ for prompt $\mathbf{x} = \{s_1, \dots, s_n\}$ and target response $\mathbf{y}$. Thus, ablating each prompt sentence $s_i$ yields its \textit{relative energy impact}:
\begin{equation}\label{eq:relative_energy_impact}
    \Delta E_i = \mathcal{E}(\mathbf{x}_{\setminus i}, \mathbf{y}) - E_{\text{base}}
\end{equation}
where $\Delta E_i > 0$ indicates that $s_i$ contributed to compatibility.

We evaluate the resulting EBMs across four diagnostic dimensions designed to probe known failure modes in models:

\vspace{0.25cm}
\noindent{1. Input Directive Dependency:} Adapting rationale extraction protocols from ERASER~\cite{deyoung2020eraser}, we evaluate whether the surrogate isolates explicit user directives ($S_Q$, sentences possessing direct interrogative pronouns or question marks) from background context ($S_R$). We define the \textit{signal-to-noise ratio}, as $\text{SNR} = (\frac{1}{|S_Q|}\sum_{i \in S_Q} \Delta E_i) / (\frac{1}{|S_R|}\sum_{k \in S_R} \Delta E_k + \epsilon)$, and the \textit{mean reciprocal rank} as $\text{MRR} = \frac{1}{N}\sum_{j=1}^N (1 / \text{rank}_{Q}^{(j)})$, which penalizes models that fail to place the directive in the primary ranking position in our sample subset ($N=769$).

\vspace{0.25cm}
\noindent{2. Output Semantic Robustness:} Text classifiers frequently succumb to spurious artifacts~\cite{gururangan2018annotation}. In instruction-tuned LLMs, conversational preambles (e.g., ``Okay!'', set $S_{\text{artifact}}$) are readily learned as superficial shortcuts. Following input reduction protocols~\cite{feng2018pathologies}, we evaluate sensitivity to non-semantic fillers on an artifact-rich subset ($N=890$). We define the \textit{artifact impact ratio}, $\text{AIR} = (\frac{1}{|S_{\text{artifact}}|}\sum_{i \in S_{\text{artifact}}} \Delta E_i) / (\frac{1}{|S_{\text{meaningful}}|}\sum_{k \in S_{\text{meaningful}}} \Delta E_k + \epsilon)$, and the \textit{rank-1 false discovery rate}, $\text{FDR@1} = \frac{1}{N}\sum_{j=1}^N \mathbb{I}[\operatorname*{argmax}_i \Delta E_{j,i} \in S_{\text{artifact},j}]$, quantifying how often a filler is erroneously assigned the highest importance.

\vspace{0.25cm}
\noindent{3. Output Information Ranking:} To scalably assess whether surrogate energy drops align with human-intelligible semantic hierarchies in extended responses, we prompt \texttt{Gemini-2.5-Flash} to score response sentences across a 6-level information density scale ($\text{rel}_i \in \{0, \dots, 5\}$, from phatic filler to critical thesis) ($N=500$). We evaluate ranking alignment via Spearman's rank correlation coefficient $\rho \in [-1, 1]$ and the \textit{normalized discounted cumulative gain} at cutoff $k=3$~\cite{jarvelin2002cumulated}, where $\text{DCG}@3 = \sum_{i=1}^3 (2^{\text{rel}_{\pi(i)}} - 1) / \log_2(i + 1)$, normalized by the ideal ranking IDCG@3.

\vspace{0.25cm}
\noindent{4. Distractor Disentanglement:} Effective attribution requires separating true antecedents from topical distractors~\cite{kaushik2020learning}. We prompt \texttt{Gemini-2.5-Flash} to generate 500 triplets $(y_{\text{target}}, x_{\text{true}}, x_{\text{distractor}})$ from validation data, where $x_{\text{distractor}}$ shares strong topical overlap with $y_{\text{target}}$ but lacks logical entailment. We define \textit{counterfactual accuracy} ($\text{Acc} = \frac{100}{N}\sum \mathbb{I}[\mathcal{E}(x_{\text{true}}, y_{\text{target}}) < \mathcal{E}(x_{\text{distractor}}, y_{\text{target}})]$) and the \textit{energy separation margin} ($\text{ESM} = \frac{1}{N}\sum [ \mathcal{E}(x_{\text{distractor}}, y_{\text{target}}) - \mathcal{E}(x_{\text{true}}, y_{\text{target}}) ]$).

\input{Tables/tab_ebm_ablation_diagnostics}

\begin{figure}[!t]
    \centering
    \subfloat[$\mathcal{E}_{\text{Fid}}$ Energy]{\includegraphics[width=0.32\columnwidth]{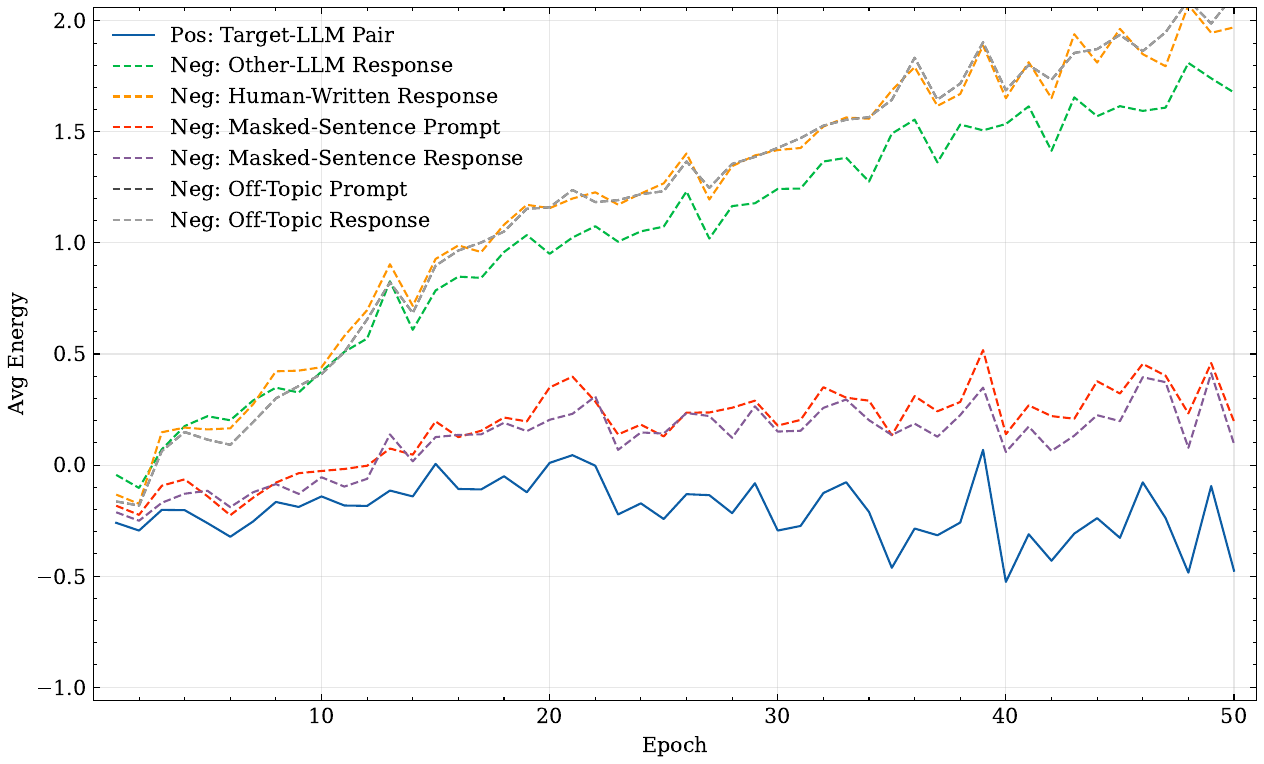}%
    \label{fig:ebm_energy_fid}}%
    \hfill
    \subfloat[$\mathcal{E}_{\text{Dep}}$ Energy]{\includegraphics[width=0.32\columnwidth]{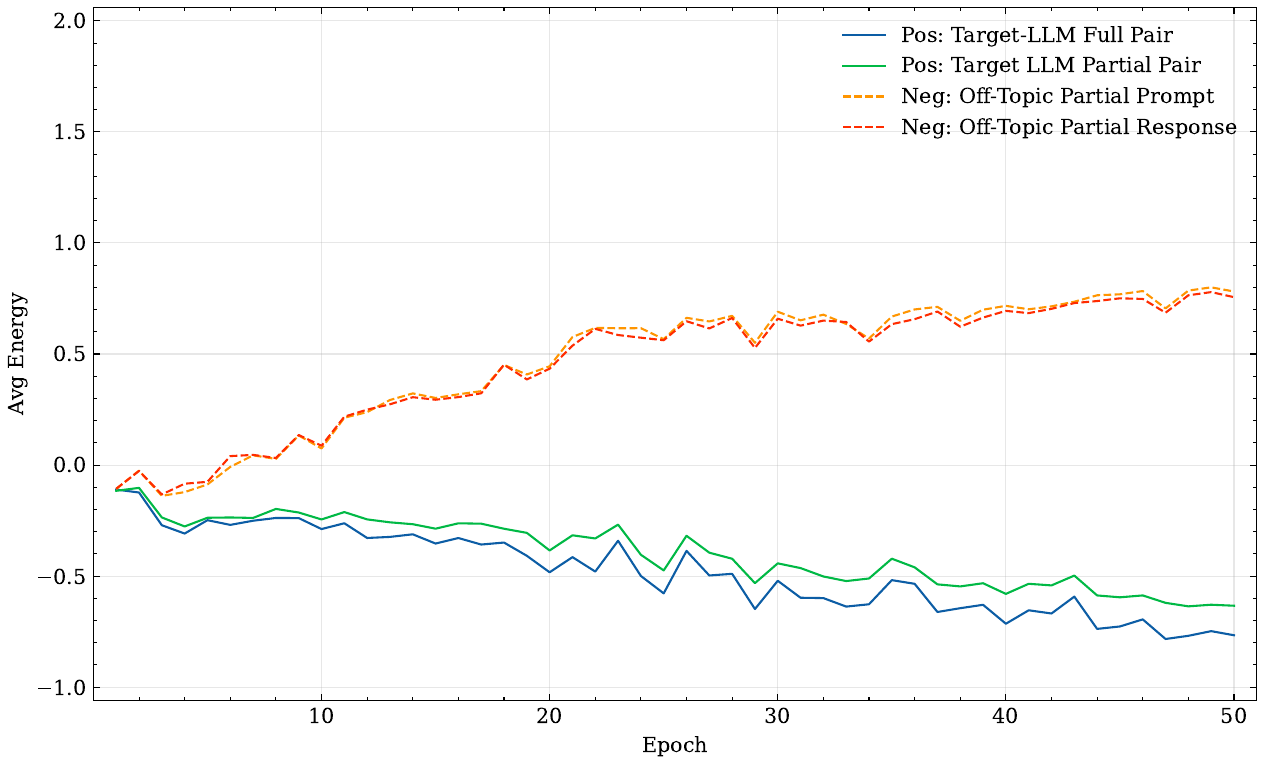}%
    \label{fig:ebm_energy_dep}}%
    \hfill
    \subfloat[$\mathcal{E}_{\text{Hyb}}$ Energy]{\includegraphics[width=0.32\columnwidth]{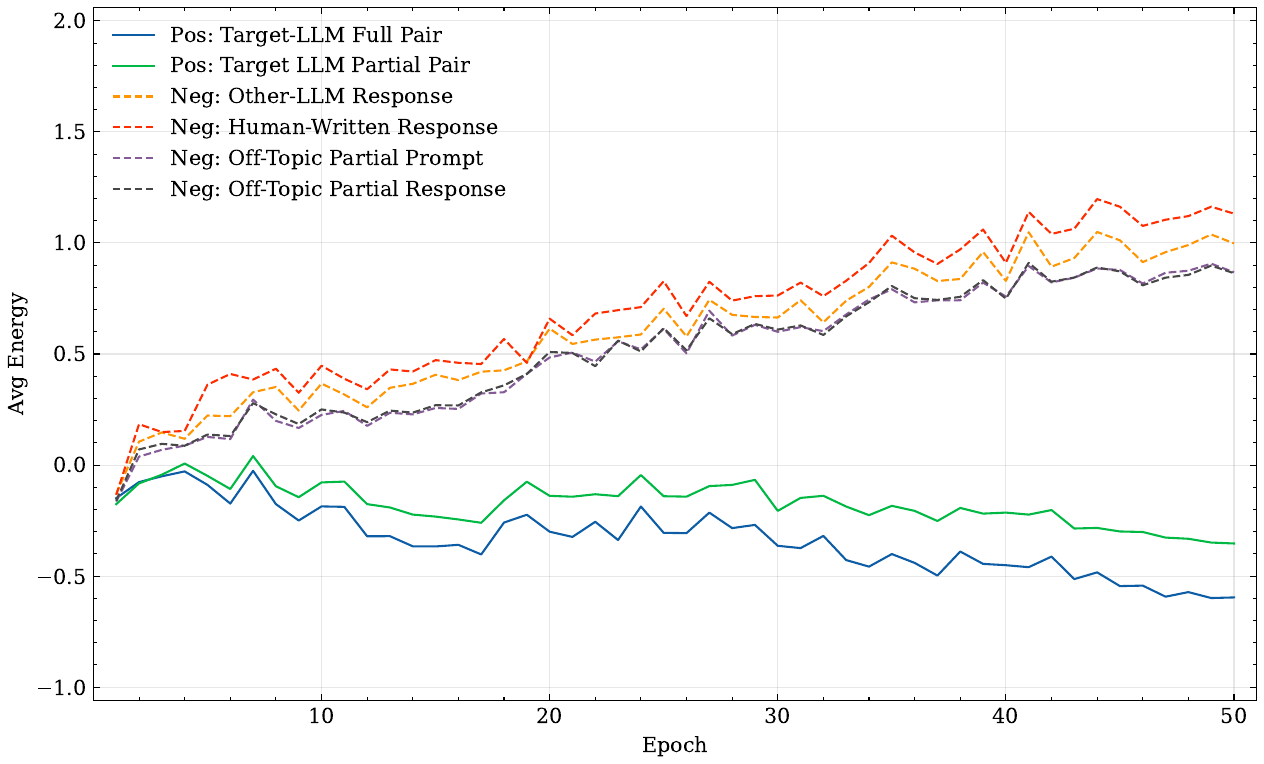}%
    \label{fig:ebm_energy_hyb}}%
    \\[2pt]
    \subfloat[$\mathcal{E}_{\text{Fid}}$ Loss]{\includegraphics[width=0.32\columnwidth]{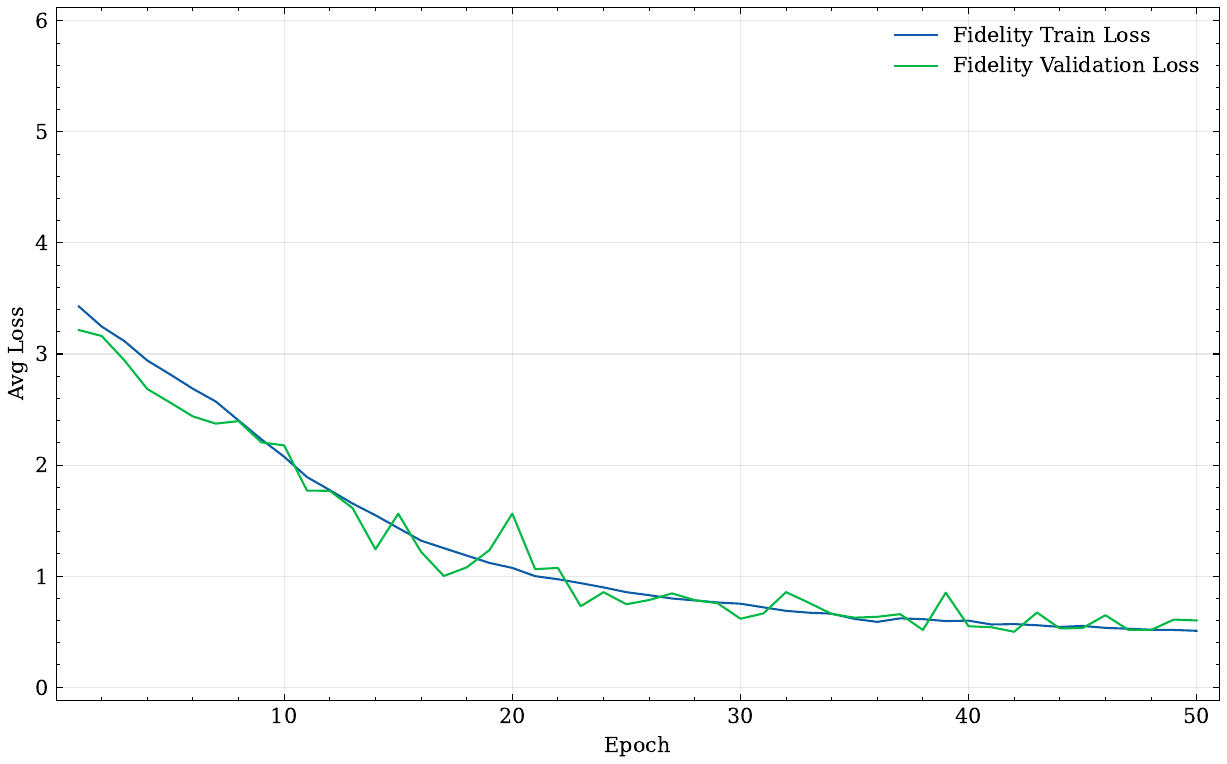}%
    \label{fig:ebm_loss_fid}}%
    \hfill
    \subfloat[$\mathcal{E}_{\text{Dep}}$ Loss]{\includegraphics[width=0.32\columnwidth]{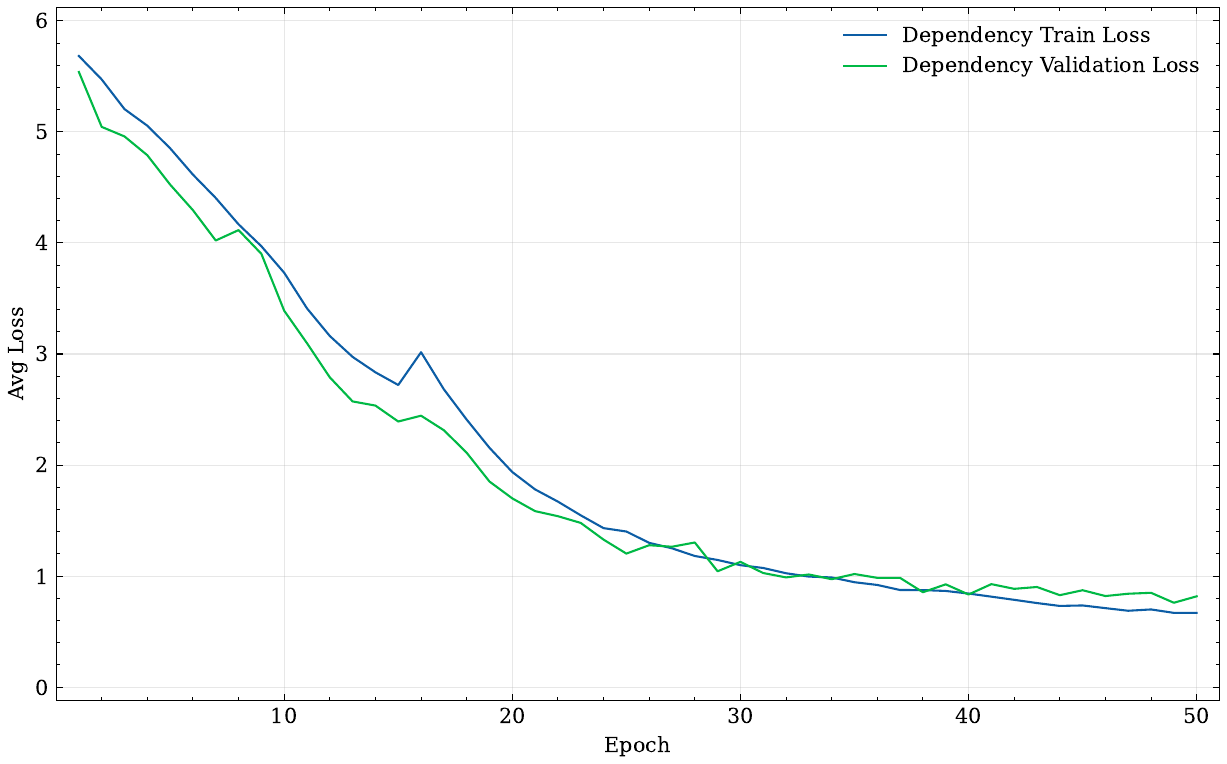}%
    \label{fig:ebm_loss_dep}}%
    \hfill
    \subfloat[$\mathcal{E}_{\text{Hyb}}$ Loss]{\includegraphics[width=0.32\columnwidth]{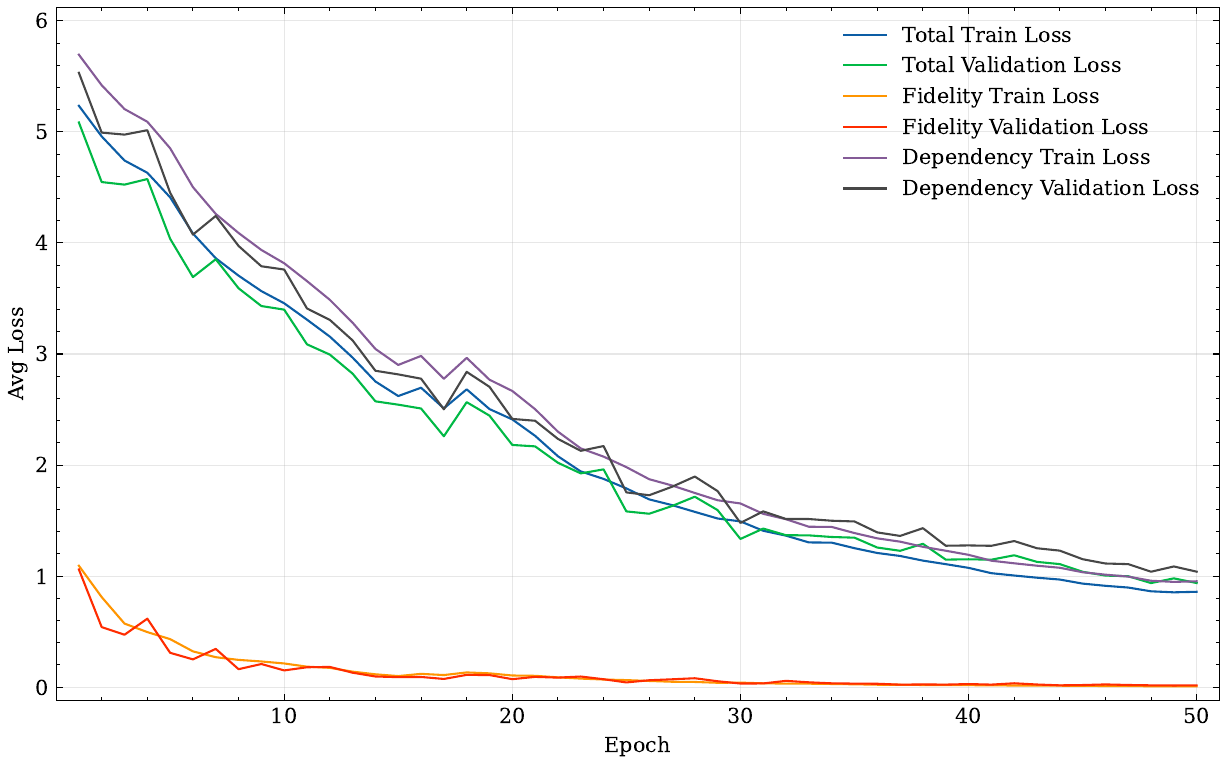}%
    \label{fig:ebm_loss_hyb}}%
    \\[2pt]
    \subfloat[$\mathcal{IN}$ under $\mathcal{E}_{\text{Dep}}$]{\includegraphics[width=0.48\columnwidth]{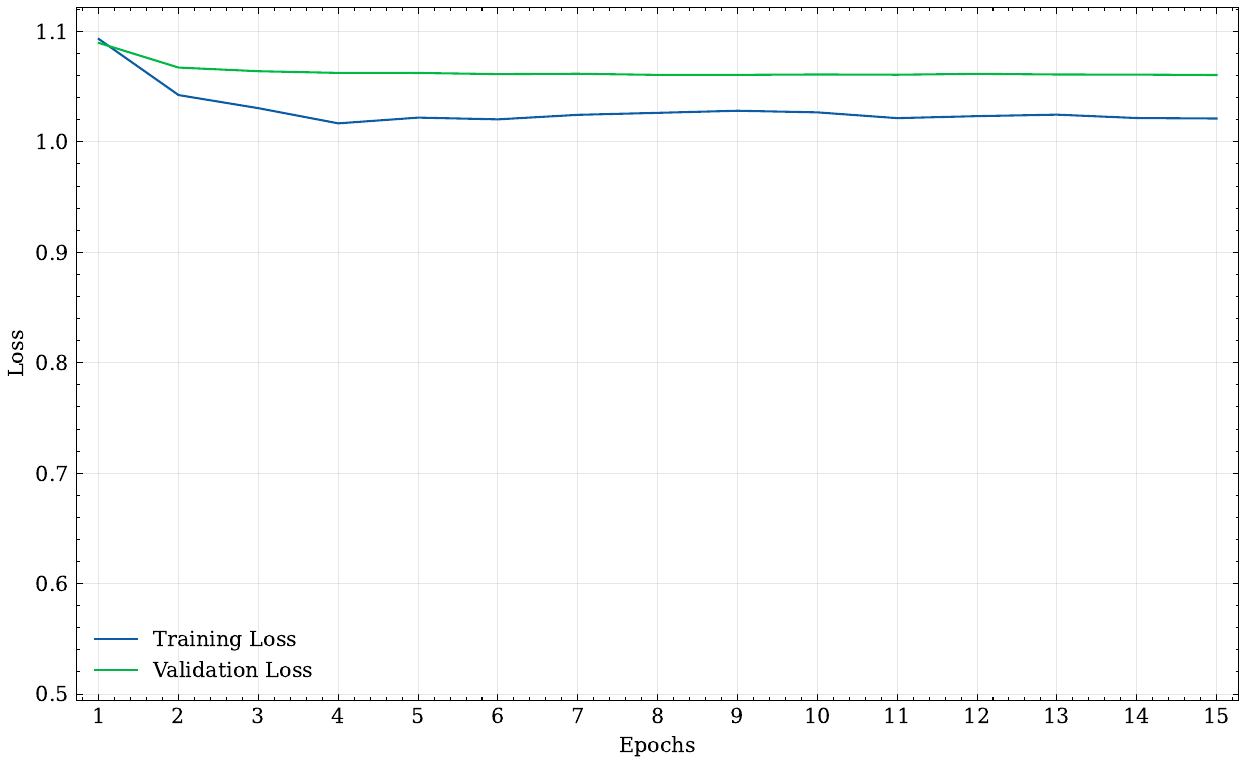}%
    \label{fig:interpreter_loss_dep}}%
    \hfill
    \subfloat[$\mathcal{IN}$ under $\mathcal{E}_{\text{Hybrid}}$]{\includegraphics[width=0.48\columnwidth]{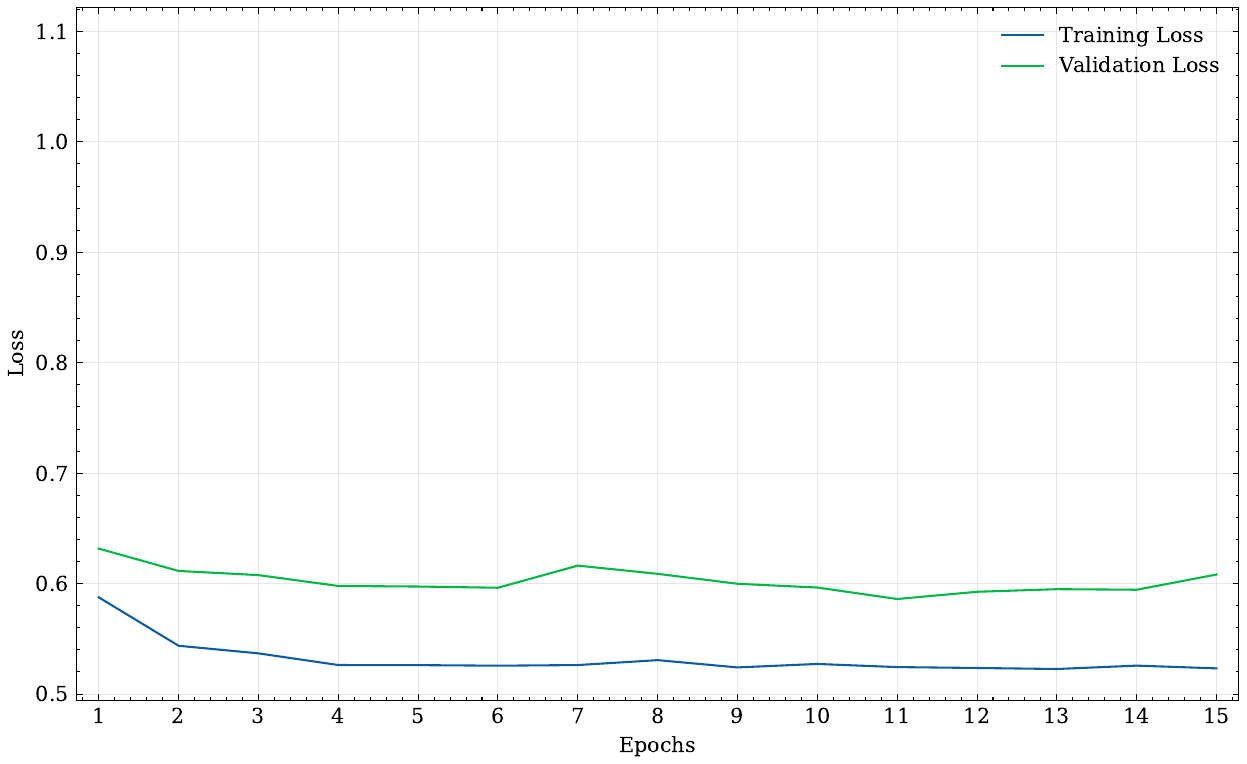}%
    \label{fig:interpreter_loss_hyb}}%
    \caption{\textbf{Surrogate pre-training and downstream interpreter optimization.} (a--c) Validation energy over 50 epochs for positive pairs (solid) vs. negative samples (dashed). (d--f) Loss curves confirming asymptotic convergence across objectives. (g--h) Downstream interpreter loss.}\label{fig:ebm_interpreter_dynamics}
\end{figure}

\vspace{0.25cm}
\paragraph{Ablation Analysis}
Table~\ref{tab:ebm_ablation_diagnostics} illustrates the failure modes of uncalibrated objectives. $\mathcal{E}_{\text{Fidelity}}$ is vulnerable to surface artifacts, frequently minimizing energy on common preambles rather than interrogative directives, which leads to low SNR and near-random counterfactual accuracy. Consequently, despite the energy margin in Fig.~\ref{fig:ebm_energy_fid}, the learned landscape reflects style discrimination rather than antecedent identification. Meanwhile, $\mathcal{E}_{\text{Dep}}$ suppresses conversational artifacts, but training exclusively on fragmented contrasts degrades paragraph-level discourse ranking ($\rho$, nDCG). This is reflected in the downstream performance; the surrogate EBM serves as the critic providing supervisory gradients to train the interpreter [$\mathcal{IN}$, (\ref{eq:interpreter_objective})] and Fig.~\ref{fig:ebm_interpreter_dynamics}(g--h) illustrates how this critic's formulation governs downstream optimization. Under $\mathcal{E}_{\text{Dep}}$, convergence fails, with validation loss plateauing (Fig.~\ref{fig:interpreter_loss_dep}); without global regularization, the energy manifold exhibits extreme sensitivity, and generates highly oscillatory and destabilizing gradients. $\mathcal{E}_{\text{Hybrid}}$ resolves this tension; local dependency samplers isolate antecedents while global fidelity preserves discourse structure, maintaining directive sensitivity while resisting conversational artifacts. Fig.~\ref{fig:interpreter_loss_hyb} shows the successful convergence of the interpreter under this hybrid objective.

\subsection{Interpreter Attribution Plausibility}\label{subsec:experiments_plausibility}

Evaluating black-box explanations is fundamentally challenging because open-ended generation lacks gold-standard attributions, while existing human annotations for simpler tasks often reflect cognitive priors rather than true model mechanics. We therefore first evaluate attribution plausibility, assessing whether the prompt sentences selected by the interpreter represent semantically and logically sound antecedents. Because high-capacity foundation models share the target LLM's autoregressive inductive biases, they provide competent proxies for open-ended text. Thus, we test on our two complementary settings of \textbf{HC3} and \textbf{WikiBio} datasets.

\vspace{0.25cm}
\paragraph{Evaluation Metrics}
For each instance, let $\mathbf{y}_{\text{pred}} \in \mathbb{R}^n$ denote the importance scores output by the interpreter for the $n$ prompt sentences, and let $\mathbf{y}_{\text{ref}} \in \mathbb{R}^n$ denote the reference distribution from an oracle or quasi-ground-truth. Attribution alignment is then quantified via two complementary metrics:

\vspace{0.25cm}
\noindent{1. Soft Top-1 Accuracy (SoftAcc).} Because multiple prompt sentences often act jointly to trigger an output sentence, strict top-1 matching is overly restrictive. SoftAcc instead tests whether the top-scoring sentence, $i^* = \operatorname*{argmax}_i {(\mathbf{y}_{\text{pred}})}_i$, falls within the top reference set $\mathcal{S}_{\text{top}} = \{i \mid {(\mathbf{y}_{\text{ref}})}_i \ge v_{\text{top2}}\}$: $\text{SoftAcc} = \mathbb{I}[i^* \in \mathcal{S}_{\text{top}}]$, where $v_{\text{top2}}$ is the second-highest unique score in $\mathbf{y}_{\text{ref}}$ to accommodate ties.

\vspace{0.25cm}
\noindent{2. Normalized Discounted Cumulative Gain (nDCG).} To assess ranking quality across the entire sequence rather than just the top candidate, we compute nDCG~\cite{jarvelin2002cumulated}. Letting $\pi$ be the permutation sorting $\mathbf{y}_{\text{pred}}$ in descending order, Discounted Cumulative Gain uses $\mathbf{y}_{\text{ref}}$ as relevance grades, given by $\text{DCG} = \sum_{j=1}^n \frac{{(\mathbf{y}_{\text{ref}})}_{\pi(j)}}{\log_2(j + 1)}$. Normalizing by the ideal ranking IDCG yields $\text{nDCG} \in [0, 1]$, penalizing interpreters that assign low importance to sentences deemed critical by the reference.

\input{Tables/tab_hc3_plausibility_blackbox}

\vspace{0.15cm}
\paragraph{Unconstrained Generation (HC3)}
Table~\ref{tab:hc3_plausibility_blackbox} presents cross-evaluation matrices on HC3 for \texttt{GPT-4o-Mini}. Despite its smaller parameter footprint, ESCI aligns with foundation model oracles in both top-1 attribution and ranking correlation (nDCG). The matrices exhibit asymmetry between rows and columns due to differing scoring behaviors; while generative oracles distribute importance across multiple context sentences, ESCI's sparsity penalty produces focused selections. Consequently, ESCI's top choices regularly fall within the oracles' top scores, whereas diffuse oracle scores align less frequently with ESCI.

\input{Tables/tab_wikibio_plausibility_blackbox}

\vspace{0.15cm}
\paragraph{Constrained Generation (WikiBio)}
On WikiBio for \texttt{Llama-3.1-8B-It} (Table~\ref{tab:wikibio_plausibility_blackbox}), ESCI aligns well with \texttt{StringMatching} and \texttt{LIME-BlackBox}, capturing semantic compatibility of antecedents in biographical synthesis. Agreement with LLM oracles is more moderate since, in structured data-to-text synthesis, larger models focus on broader entity tracking.

\vspace{0.15cm}
\paragraph{White-Box and Gray-Box Plausibility}
Table~\ref{tab:wikibio_plausibility_whitebox} compares ESCI on WikiBio against white-box and gray-box baselines. Without internal model access, ESCI shows comparable alignment to attention-based methods (\texttt{AttentionRollout}) and \texttt{StringMatching}, whereas gradient-based methods display lower consensus due to token-level gradient dispersion.

\input{Tables/tab_wikibio_plausibility_whitebox}

\vspace{-0.25cm}
\subsection{Interpreter Faithfulness}\label{subsec:experiments_faithfulness}

While plausibility evaluates semantic alignment with references, it does not confirm whether the selected sentences influence the LLM's actual generation. We evaluate attribution faithfulness by measuring output changes under input perturbations~\cite{atanasova2023faithfulness}. In generative settings, such interventions test whether identified antecedents functionally alter the output, providing an estimate of behavioral influence. Following rationale evaluation frameworks~\cite{deyoung2020eraser}, we adapt \textit{Sufficiency} and \textit{Comprehensiveness} to sentence-level generation. \textbf{Generative Sufficiency} ($\mathcal{S}$) measures retention of target $\mathbf{y}_T$ when the model is prompted with only the selected sentences $\mathbf{x}_S$. \textbf{Generative Comprehensiveness} ($\mathcal{C}$) measures degradation when $\mathbf{x}_S$ is removed. A faithful interpreter aims for high sufficiency and low comprehensiveness, yielding a positive \textbf{Faithfulness Gap}:
\begin{equation}\label{eq:faithfulness_gap}
    \resizebox{0.95\linewidth}{!}{$
        \mathcal{S} = \mathcal{F}\big(\text{LLM}(\mathbf{x}_S), \mathbf{y}_T\big), \;\; 
        \mathcal{C} = \mathcal{F}\big(\text{LLM}(\mathbf{x} \setminus \mathbf{x}_S), \mathbf{y}_T\big), \;\; 
        \text{Gap} = \mathcal{S} - \mathcal{C}
    $}
\end{equation}
where $\mathcal{F}$ evaluates retention of target $\mathbf{y}_T$ and is defined below.

\vspace{0.15cm}
\paragraph{Evaluation Metrics}
Generative models possess high linguistic fluidity. To provide a thorough assessment, we instantiate $\mathcal{F}$ in three ways:

\vspace{0.25cm}
\noindent{1. NLI Logical Entailment ($\mathcal{F}_{\text{NLI}}$):} As our primary gold standard, NLI tests whether the newly generated full text entails the previously chosen target sentence, overcoming both lexical rigidity and information dispersion across multiple sentences. We feed the new text $\tilde{\mathbf{y}}$ as the premise and target sentence $\mathbf{y}_T$ as the hypothesis into \texttt{cross-encoder/nli-deberta-v3-large}~\cite{he2023debertav}. Zeroing the neutral class forces a binary decision between entailment ($z_{\text{ent}}$) and contradiction ($z_{\text{con}}$):
\begin{equation}\label{eq:metric_nli}
    \mathcal{F}_{\text{NLI}}(\tilde{\mathbf{y}}, \mathbf{y}_T) = \frac{\exp(z_{\text{ent}})}{\exp(z_{\text{ent}}) + \exp(z_{\text{con}})}
\end{equation}

\vspace{0.1cm}
\noindent{2. Maximum Sentence-Level Cosine Similarity ($\mathcal{F}_{\text{cosine}}$):} We compute semantic similarity between the new text and the target sentence using \texttt{all-mpnet-base-v2}~\cite{song2020mpnet}. To prevent embedding dilution from compressing a paragraph into a single vector, we segment the new text $\tilde{\mathbf{y}}$ into sentences $\{\tilde{y}_1, \dots, \tilde{y}_K\}$ and record the maximum cosine similarity to $\mathbf{y}_T$:
\begin{equation}\label{eq:metric_cosine}
    \mathcal{F}_{\text{cosine}}(\tilde{\mathbf{y}}, \mathbf{y}_T) = \max_{i \in \{1, \dots, K\}} \cos\big(e(\tilde{y}_i), e(\mathbf{y}_T)\big)
\end{equation}

\vspace{0.1cm}
\noindent{3. Normalized Log-Likelihood ($\mathcal{F}_{\text{logprob}}$):} We also evaluate forced-decoding token probabilities over the $M$ tokens of $\mathbf{y}_T$: $\mathcal{L}(\mathbf{x}', \mathbf{y}_T) = \sum_{j=1}^M \log P(y_j \mid y_{<j}, \mathbf{x}')$. To enable bounded comparison, log-likelihoods are scaled relative to the prompt $\mathcal{L}(\mathbf{x}, \mathbf{y}_T)$ and empty context $\mathcal{L}(\emptyset, \mathbf{y}_T)$ with min-max clamping:
\begin{equation}\label{eq:metric_logprob}
    \mathcal{F}_{\text{logprob}}(\mathbf{x}', \mathbf{y}_T) = {\left[ \frac{\mathcal{L}(\mathbf{x}', \mathbf{y}_T) - \mathcal{L}(\emptyset, \mathbf{y}_T)}{\mathcal{L}(\mathbf{x}, \mathbf{y}_T) - \mathcal{L}(\emptyset, \mathbf{y}_T) + \epsilon} \right]}_0^1
\end{equation}

All generative evaluations ($\mathcal{F}_{\text{NLI}}$, $\mathcal{F}_{\text{cosine}}$) are averaged across 3 seeds due to containing randomness in their computation.

\input{Tables/tab_wikibio_faithfulness_blackbox}

\vspace{0.25cm}
\paragraph{Black-Box Analysis}
Table~\ref{tab:wikibio_faithfulness_blackbox}(a) reports faithfulness under NLI entailment. Random sentence selection yields negative gaps, showing that arbitrary subsets degrade output. In contrast, ESCI achieves positive gaps that exceed target self-explanations and approach oracle explainers. Across all methods, comprehensiveness remains bounded, as models can generate plausible text from parametric memory~\cite{kamahi2024counterfactuals}.

Under sentence cosine similarity (Table~\ref{tab:wikibio_faithfulness_blackbox}(b)), self-explanations yield negative gaps, whereas ESCI maintains positive gaps comparable to oracles. Lexical methods (\texttt{StringMatching} and \texttt{LIME-BlackBox}) record higher cosine gaps due to direct surface token overlap.

Finally, under normalized log-likelihood (Table~\ref{tab:wikibio_faithfulness_blackbox}(c)), \texttt{StringMatching} and \texttt{LIME-BlackBox} maintain an advantage and preserve next-token probabilities due to verbatim phrasing, yet ESCI achieves positive retention gaps without logit access, outperforming target self-attributions.

\input{Tables/tab_wikibio_faithfulness_whitebox}

\vspace{0.15cm}
\paragraph{White-Box and Gray-Box Analysis}
Table~\ref{tab:wikibio_faithfulness_whitebox} compares faithfulness against white-box and gray-box baselines with access to internal activations or logits. Here, we evaluate exclusively under our gold-standard NLI entailment metric ($\mathcal{F}_{\text{NLI}}$), as token-level metrics suffer from circular evaluation---granting an unfair advantage to methods directly optimizing or scoring the very token log-likelihoods or attention representations they are tested on. Under this semantic standard, ESCI achieves positive faithfulness gaps and remains competitive with privileged-access baselines without requiring internal model access.

\input{Tables/tab_hc3_attribution_qualitative}

\vspace{0.15cm}
\paragraph{Sample Analysis}
Table~\ref{tab:hc3_attribution_qualitative} presents representative samples selected by clustering prompt-response pairs via \textit{total variation distance} against oracle distributions into consensus, divergence, and ambiguity. Across these regimes, ESCI consistently mirrors frontier oracles on direct semantic mappings and effectively disentangles stylistic constraints from topical content, whereas complex multi-sentence synthesis naturally leads to distributed attribution across all evaluators.

\subsection{Computational Scalability}\label{subsec:experiments_scalability}

Practical deployment requires balancing attribution fidelity against query latency and computational overhead. As shown in Table~\ref{tab:model_scale_comparison}, continuous auditing with frontier LLM oracles incurs unsustainable API costs, while smaller open-weights models ($<10$B) fail to reliably output calibrated probability vectors without extensive prompting. Conversely, perturbation baselines like \texttt{LIME} face severe combinatorial bottlenecks.

\begin{figure}[h]
    \centering
    \includegraphics[width=\columnwidth]{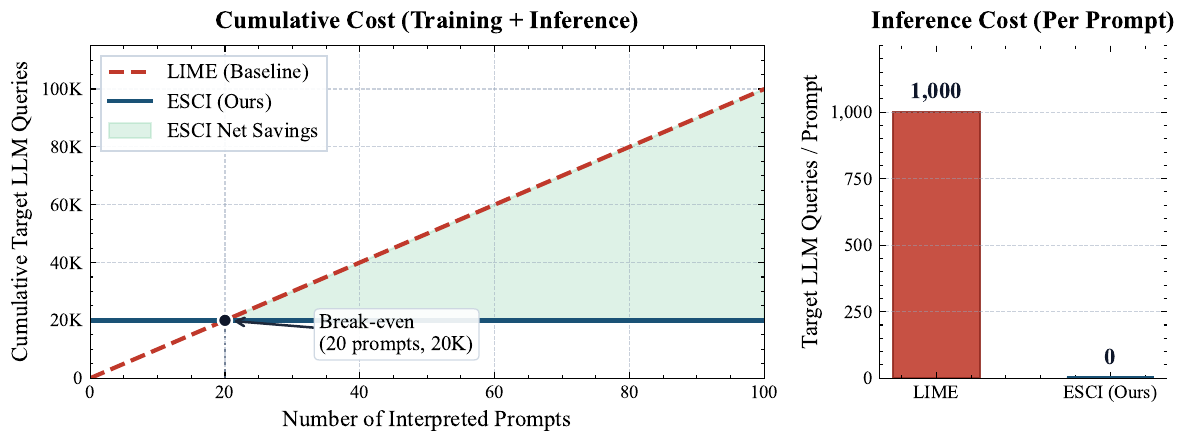}
    \caption{\textbf{Computational Scalability Comparison.} \textbf{(Left)} Cumulative target LLM API calls (training + inference) scaling with interpreted prompts. \textbf{(Right)} Per-instance inference API queries, demonstrating ESCI's amortized efficiency.}\label{fig:computational_scalability}
\end{figure}

For a prompt of $N$ sentences, \texttt{LIME} requires $\min(2^N, K_{\text{max}})$ queries (capped at $K_{\text{max}} = 1000$). Because benchmark prompts average $N \approx 15$ sentences, this ceiling is consistently saturated, requiring $1{,}000$ API queries per prompt. In contrast, ESCI shifts computation to offline training ($20{,}000$ queries). Once trained, it operates in $\mathcal{O}(1)$ time with zero additional API queries to the target LLM. Accounting for upfront data collection, ESCI breaks even with \texttt{LIME} after auditing only $20$ prompts, incurring no recurring query overhead thereafter (Fig.~\ref{fig:computational_scalability}).

\section{Conclusion and Future Work}\label{sec:conclusion}

We introduced a post-hoc attribution framework that interprets black-box language models at the sentence level. By learning a surrogate energy landscape that captures both macro-level consistency and local dependencies, our amortized interpreter identifies influential prompt antecedents in $\mathcal{O}(1)$ time without inference-time API queries. The framework is inherently modular, and can be trained with modest data to interpret specialized, single-task applications, while naturally scaling to larger surrogate capacities and multi-task datasets as more data becomes available. Additionally, despite being an attribution of observed outputs, the identified antecedents offer evidence for potential causal relationships in generation. Viewing black-box attribution through this lens provides a practical foundation for future interpretability research.

Several limitations point to promising directions for future work. First, evaluating attribution faithfulness in natural language generation is inherently constrained by the lack of deterministic ground truth, as internal autoregressive pathways are unobservable in black-box models. Second, while ESCI achieves $\mathcal{O}(1)$ inference, it shifts computation to offline pre-training, and the evaluated surrogates remain task-specific. Scaling surrogate architectures to larger multi-task corpora will be necessary to achieve out-of-the-box generalization over complex, long-context prompts. Finally, extending sentence-level attributions to downstream applications---such as prompt compression to reduce inference latency, chain-of-thought verification, and practical hallucination or safety auditing---presents a promising direction for future research.

\section*{Acknowledgment}

This work was supported in part by the Department of Mathematical Sciences at Sharif University of Technology through institutional research support. The authors gratefully acknowledge Kaggle for providing free cloud computing resources equipped with dual NVIDIA Tesla T4 GPUs ($2\times 16$\,GB). During the preparation of this manuscript, the authors used Google Gemini (specifically Gemini 3.8 Flash) to assist in refining language fluency and conciseness. The authors edited all content and take full responsibility for the manuscript.

\bibliographystyle{IEEEtran}
\bibliography{references}

\vspace{7cm}

\vspace{-2\baselineskip}
\begin{IEEEbiographynophoto}{Maryam Rezaee}
received the M.S. degree in computer science from Sharif University of Technology, Tehran, Iran. Her research interests include explainable artificial intelligence, reasoning in large language models, cognitively-inspired AI, and neurosymbolic architectures, driven by a vision of democratic and access­ible AI that is verifi­able and fundamentally aligned with human thought.
\end{IEEEbiographynophoto}
\vspace{-2\baselineskip}

\begin{IEEEbiographynophoto}{Pooriya Safaei}
received the B.S. degree in computer science from Sharif University of Technology, Tehran, Iran, where he is currently pursuing his M.S. degree. His research interests include trustworthy artificial intelligence, model interpretability, reasoning faithfulness, and System 2 AI, with a particular focus on large language and vision-language models.
\end{IEEEbiographynophoto}
\vspace{-2\baselineskip}

\begin{IEEEbiographynophoto}{Maryam Asgarinezhad}
received the B.S. degree in computer science from Sharif University of Technology, Tehran, Iran. She is currently an M.S. student in Computing Science at the University of Alberta. Her research interests span explainability and provenance in large language models, with a focus on identifying what drives a model's outputs.
\end{IEEEbiographynophoto}
\vspace{-2\baselineskip}

\begin{IEEEbiographynophoto}{Fatemeh Seyyedsalehi}
received the Ph.D. degree from Sharif University of Technology, Tehran, Iran, and was previously a visiting Ph.D. researcher at the University of California, Berkeley, USA. She is currently an Assistant Professor in Computer Science at Sharif University of Technology. Her research interest include deep generative models, trustworthy and interpretable machine learning, and Bayesian modeling.
\end{IEEEbiographynophoto}

\vfill

\newpage

\appendices
\input{appendix}

\end{document}

%% file: Tables/tab_model_scale_comparison.tex
\begin{table}[h]
    \centering
    \scriptsize
    \caption{\textbf{Model Scale Comparison.} Parameter footprints of our framework alongside baselines.}\label{tab:model_scale_comparison}\label{tab_model_scale_comparison}
    \setlength{\tabcolsep}{4.5pt}
    \renewcommand{\arraystretch}{1.1}
    \begin{tabularx}{\columnwidth}{l c X}
        \toprule
        \textbf{Model} & \textbf{Access} & \textbf{Parameters (Approx.)} \\
        \midrule
        \rowcolor{gray!20} \textbf{ESCI (Ours)} & Local Surrogate & \textbf{181M} ($\sim$71M Trainable) \\
        Mistral-7B-Instruct-v0.2 & Open-Weights & 7B \\
        Llama-3.1-8B-Instruct & Open-Weights & 8B \\
        Gemma-2-9B-It & Open-Weights & 9B \\
        GPT-4o-Mini & Proprietary API & Undisclosed \\
        Gemini-2.5-Flash & Proprietary API & Undisclosed \\
        \bottomrule
    \end{tabularx}
\end{table}

%% file: Tables/tab_ebm_ablation_diagnostics.tex
\begin{table}[!t]
    \centering
    \scriptsize
    \caption{\textbf{EBM Ablation Results Across Diagnostic Dimensions.} Evaluated on \texttt{GPT-4o-Mini} generations from HC3.}\label{tab:ebm_ablation_diagnostics}\label{tab_ebm_ablation_diagnostics}
    \setlength{\tabcolsep}{4.0pt}
    \renewcommand{\arraystretch}{1.15}
    \begin{tabular}{l|cc|cc|cc|cc}
        \toprule
        & \multicolumn{2}{c|}{\textbf{I. Directive}} & \multicolumn{2}{c|}{\textbf{II. Robust.}} & \multicolumn{2}{c|}{\textbf{III. Ranking}} & \multicolumn{2}{c}{\textbf{IV. Disentang.}} \\
        \textbf{Model} & \textbf{SNR}$\uparrow$ & \textbf{MRR}$\uparrow$ & \textbf{AIR}$\downarrow$ & \textbf{FDR}$\downarrow$ & \textbf{$\rho$}$\uparrow$ & \textbf{nDCG}$\uparrow$ & \textbf{ESM}$\uparrow$ & \textbf{Acc}$\uparrow$ \\
        \midrule
        $\mathcal{M}_{\text{Fidelity}}$ & 1.45 & 0.685 & 0.743 & 85.1\% & 0.231 & 0.553 & 0.125 & 57.2\% \\
        $\mathcal{M}_{\text{Dep}}$ & 6.89 & 0.822 & \textbf{0.000} & \textbf{6.9\%} & 0.485 & 0.699 & \textbf{0.499} & 84.2\% \\
        \rowcolor{gray!20} $\mathcal{M}_{\text{Hybrid}}$ & \textbf{7.15} & \textbf{0.894} & 0.173 & 15.2\% & \textbf{0.712} & \textbf{0.836} & 0.382 & \textbf{95.4\%} \\
        \bottomrule
    \end{tabular}
\end{table}

%% file: Tables/tab_hc3_plausibility_blackbox.tex
\begin{table}[!t]
    \centering
    \caption{\textbf{Attribution Plausibility Matrices on HC3 (Black-Box).} Interpreter in each Row $i$ is evaluated against Reference Column $j$.}\label{tab:hc3_plausibility_blackbox}\label{tab_hc3_plausibility_blackbox}
    \resizebox{\columnwidth}{!}{%
    \setlength{\tabcolsep}{3.5pt}
    \renewcommand{\arraystretch}{1.1}
    \begin{tabular}{l|c|ccccc}
        \toprule
        \multicolumn{7}{c}{\textbf{(a) Soft Top-1 Accuracy}} \\
        \textbf{Interp. $\downarrow$ $\mid$ Ref. $\rightarrow$} & \textbf{Ours} & \textbf{Gemini} & \textbf{4o-Mini} & \textbf{Gemma} & \textbf{Llama} & \textbf{Mistral} \\
        \midrule
        \rowcolor{gray!20} \textbf{ESCI (Ours)} & \textbf{1.000} & 0.750 & 0.640 & 0.700 & 0.720 & 0.740 \\
        Gemini-2.5-Flash                & 0.670 & \textbf{1.000} & \textbf{0.750} & 0.650 & 0.680 & 0.670 \\
        GPT-4o-Mini                     & 0.600 & \textbf{0.850} & \textbf{1.000} & 0.620 & 0.640 & 0.650 \\
        Gemma-2-9B-It                   & 0.690 & 0.650 & 0.490 & \textbf{1.000} & \textbf{0.830} & 0.780 \\
        Llama-3.1-8B-It                 & \textbf{0.720} & 0.680 & 0.510 & \textbf{0.850} & \textbf{1.000} & \textbf{0.800} \\
        Mistral-7B-It                   & 0.700 & 0.670 & 0.520 & 0.790 & 0.810 & \textbf{1.000} \\
        \midrule
        \multicolumn{7}{c}{\textbf{(b) Normalized Discounted Cumulative Gain (nDCG)}} \\
        \textbf{Interp. $\downarrow$ $\mid$ Ref. $\rightarrow$} & \textbf{Ours} & \textbf{Gemini} & \textbf{4o-Mini} & \textbf{Gemma} & \textbf{Llama} & \textbf{Mistral} \\
        \midrule
        \rowcolor{gray!20} \textbf{ESCI (Ours)} & \textbf{1.000} & 0.830 & 0.790 & 0.810 & 0.820 & 0.830 \\
        Gemini-2.5-Flash                & \textbf{0.850} & \textbf{1.000} & \textbf{0.880} & 0.760 & 0.780 & 0.790 \\
        GPT-4o-Mini                     & 0.770 & \textbf{0.890} & \textbf{1.000} & 0.770 & 0.790 & 0.800 \\
        Gemma-2-9B-It                   & 0.800 & 0.750 & 0.740 & \textbf{1.000} & \textbf{0.860} & 0.830 \\
        Llama-3.1-8B-It                 & 0.820 & 0.770 & 0.750 & \textbf{0.880} & \textbf{1.000} & \textbf{0.850} \\
        Mistral-7B-It                   & 0.810 & 0.780 & 0.760 & 0.820 & 0.840 & \textbf{1.000} \\
        \bottomrule
    \end{tabular}%
    }
\end{table}

%% file: Tables/tab_wikibio_plausibility_blackbox.tex
\begin{table}[!t]
    \centering
    \caption{\textbf{Attribution Plausibility Matrices on WikiBio (Black-Box).} Interpreter in each Row $i$ is evaluated against Reference Column $j$.}\label{tab:wikibio_plausibility_blackbox}\label{tab_wikibio_plausibility_blackbox}
    \resizebox{\columnwidth}{!}{%
    \setlength{\tabcolsep}{2.5pt}
    \renewcommand{\arraystretch}{1.1}
    \begin{tabular}{l|c|cccccc}
        \toprule
        \multicolumn{8}{c}{\textbf{(a) Soft Top-1 Accuracy}} \\
        \textbf{Interp. $\downarrow$ $\mid$ Ref. $\rightarrow$} & \textbf{Ours} & \textbf{Str.} & \textbf{LIME-Blk} & \textbf{Gemini} & \textbf{Gemma} & \textbf{Llama} & \textbf{Mist.} \\
        \midrule
        \rowcolor{gray!20} \textbf{ESCI (Ours)} & \textbf{1.000} & 0.712 & 0.698 & 0.591 & 0.425 & 0.439 & 0.496 \\
        \rowcolor{gray!8} StringMatching & \textbf{0.991} & \textbf{1.000} & \textbf{1.000} & \textbf{0.880} & \textbf{0.992} & \textbf{0.956} & \textbf{0.981} \\
        LIME-BlackBox                   & 0.395 & \textbf{0.944} & \textbf{1.000} & 0.670 & 0.524 & 0.533 & 0.422 \\
        Gemini-2.5-Flash                & 0.359 & 0.620 & 0.610 & \textbf{1.000} & 0.450 & 0.460 & 0.440 \\
        Gemma-2-9B-It                   & 0.362 & 0.251 & 0.369 & 0.470 & \textbf{1.000} & 0.791 & 0.637 \\
        Llama-3.1-8B-It                 & 0.357 & 0.321 & 0.426 & 0.480 & 0.807 & \textbf{1.000} & 0.640 \\
        Mistral-7B-It                   & 0.485 & 0.337 & 0.421 & 0.450 & 0.679 & 0.675 & \textbf{1.000} \\
        \midrule
        \multicolumn{8}{c}{\textbf{(b) Normalized Discounted Cumulative Gain (nDCG)}} \\
        \textbf{Interp. $\downarrow$ $\mid$ Ref. $\rightarrow$} & \textbf{Ours} & \textbf{Str.} & \textbf{LIME-Blk} & \textbf{Gemini} & \textbf{Gemma} & \textbf{Llama} & \textbf{Mist.} \\
        \midrule
        \rowcolor{gray!20} \textbf{ESCI (Ours)} & \textbf{1.000} & 0.637 & 0.658 & 0.621 & 0.562 & 0.569 & 0.615 \\
        \rowcolor{gray!8} StringMatching & \textbf{0.741} & \textbf{1.000} & \textbf{0.915} & \textbf{0.860} & \textbf{0.811} & \textbf{0.819} & \textbf{0.765} \\
        LIME-BlackBox                   & 0.646 & 0.697 & \textbf{1.000} & 0.660 & 0.660 & 0.676 & 0.627 \\
        Gemini-2.5-Flash                & 0.577 & \textbf{0.701} & 0.687 & \textbf{1.000} & 0.550 & 0.560 & 0.540 \\
        Gemma-2-9B-It                   & 0.493 & 0.589 & 0.652 & 0.560 & \textbf{1.000} & 0.805 & 0.716 \\
        Llama-3.1-8B-It                 & 0.478 & 0.565 & 0.631 & 0.570 & 0.768 & \textbf{1.000} & 0.673 \\
        Mistral-7B-It                   & 0.579 & 0.586 & 0.646 & 0.550 & 0.750 & 0.741 & \textbf{1.000} \\
        \bottomrule
    \end{tabular}%
    }
\end{table}

%% file: Tables/tab_wikibio_plausibility_whitebox.tex
\begin{table}[h]
    \centering
    \caption{\textbf{Attribution Plausibility Matrices on WikiBio (White-Box and Gray-Box).} Interpreter in each Row $i$ is evaluated against Reference Column $j$.}\label{tab:wikibio_plausibility_whitebox}\label{tab_wikibio_plausibility_whitebox}
    \resizebox{\columnwidth}{!}{%
    \setlength{\tabcolsep}{2.5pt}
    \renewcommand{\arraystretch}{1.1}
    \begin{tabular}{l|c|ccccccc}
        \toprule
        \multicolumn{9}{c}{\textbf{(a) Soft Top-1 Accuracy}} \\
        \textbf{Interp. $\downarrow$ $\mid$ Ref. $\rightarrow$} & \textbf{Ours} & \textbf{Str.} & \textbf{Attn.} & \textbf{Roll.} & \textbf{G-SHAP} & \textbf{Inp$\times$G} & \textbf{S-LOO} & \textbf{LIME} \\
        \midrule
        \rowcolor{gray!20} \textbf{ESCI (Ours)} & \textbf{1.000} & 0.712 & 0.613 & 0.712 & 0.461 & 0.281 & 0.581 & 0.471 \\
        \rowcolor{gray!8} StringMatching & \textbf{0.991} & \textbf{1.000} & \textbf{1.000} & \textbf{1.000} & \textbf{1.000} & \textbf{1.000} & \textbf{1.000} & \textbf{1.000} \\
        Attention                       & 0.665 & 0.905 & \textbf{1.000} & 0.909 & 0.467 & 0.286 & 0.432 & 0.373 \\
        AttentionRollout                & 0.615 & \textbf{0.996} & 0.755 & \textbf{1.000} & 0.403 & 0.310 & 0.119 & 0.118 \\
        GradientSHAP                    & 0.388 & 0.346 & 0.424 & 0.349 & \textbf{1.000} & 0.281 & 0.488 & 0.401 \\
        Input$\times$Gradient          & 0.195 & 0.159 & 0.229 & 0.161 & 0.258 & \textbf{1.000} & 0.251 & 0.196 \\
        SentenceLOO                     & 0.263 & 0.128 & 0.319 & 0.129 & 0.409 & 0.259 & \textbf{1.000} & 0.631 \\
        LIME-GrayBox                    & 0.364 & 0.168 & 0.355 & 0.167 & 0.371 & 0.205 & 0.761 & \textbf{1.000} \\
        \midrule
        \multicolumn{9}{c}{\textbf{(b) Normalized Discounted Cumulative Gain (nDCG)}} \\
        \textbf{Interp. $\downarrow$ $\mid$ Ref. $\rightarrow$} & \textbf{Ours} & \textbf{Str.} & \textbf{Attn.} & \textbf{Roll.} & \textbf{G-SHAP} & \textbf{Inp$\times$G} & \textbf{S-LOO} & \textbf{LIME} \\
        \midrule
        \rowcolor{gray!20} \textbf{ESCI (Ours)} & \textbf{1.000} & 0.637 & 0.743 & 0.726 & 0.607 & 0.508 & 0.625 & 0.624 \\
        \rowcolor{gray!8} StringMatching & 0.741 & \textbf{1.000} & \textbf{0.901} & 0.826 & 0.787 & 0.713 & \textbf{0.864} & \textbf{0.930} \\
        Attention                       & \textbf{0.760} & \textbf{0.859} & \textbf{1.000} & \textbf{0.929} & \textbf{0.860} & \textbf{0.805} & 0.863 & 0.863 \\
        AttentionRollout                & 0.710 & 0.642 & 0.882 & \textbf{1.000} & 0.631 & 0.538 & 0.525 & 0.544 \\
        GradientSHAP                    & 0.581 & 0.635 & 0.722 & 0.697 & \textbf{1.000} & 0.627 & 0.722 & 0.693 \\
        Input$\times$Gradient          & 0.494 & 0.584 & 0.618 & 0.614 & 0.630 & \textbf{1.000} & 0.626 & 0.609 \\
        SentenceLOO                     & 0.584 & 0.647 & 0.679 & 0.530 & 0.696 & 0.568 & \textbf{1.000} & 0.825 \\
        LIME-GrayBox                    & 0.598 & 0.724 & 0.705 & 0.557 & 0.691 & 0.578 & 0.837 & \textbf{1.000} \\
        \bottomrule
    \end{tabular}%
    }
\end{table}

%% file: Tables/tab_wikibio_faithfulness_blackbox.tex
\begin{table}[!t]
    \centering
    \scriptsize
    \caption{\textbf{Faithfulness Evaluation on WikiBio (Black-Box).} Mean and standard deviation across random seeds under three complementary evaluation functions.}\label{tab:wikibio_faithfulness_blackbox}\label{tab_wikibio_faithfulness_blackbox}
    \resizebox{\columnwidth}{!}{%
    \setlength{\tabcolsep}{2.5pt}
    \renewcommand{\arraystretch}{1.08}
    \begin{tabular}{l|ccc|ccc}
        \toprule
        & \multicolumn{3}{c|}{\textbf{Llama-3.1-8B-Instruct}} & \multicolumn{3}{c}{\textbf{Gemma-2-9B-It}} \\
        \textbf{Method} & \textbf{Suff.}$\uparrow$ & \textbf{Comp.}$\downarrow$ & \textbf{Gap}$\uparrow$ & \textbf{Suff.}$\uparrow$ & \textbf{Comp.}$\downarrow$ & \textbf{Gap}$\uparrow$ \\
        \midrule
        \multicolumn{7}{c}{\textbf{(a) NLI Logical Entailment ($\mathcal{F}_{\text{NLI}}$)}} \\
        \midrule
        \rowcolor{gray!20} \textbf{ESCI (Ours)} & 0.620\std{0.002} & 0.512\std{0.003} & 0.108\std{0.004} & 0.800\std{0.003} & 0.572\std{0.003} & 0.228\std{0.004} \\
        \rowcolor{gray!8} StringMatching & 0.602\std{0.004} & 0.542\std{0.005} & 0.060\std{0.004} & 0.782\std{0.004} & 0.602\std{0.005} & 0.180\std{0.006} \\
        \rowcolor{gray!8} Random (Dir.)  & 0.408\std{0.004} & 0.652\std{0.006} & --0.244\std{0.009} & 0.531\std{0.004} & 0.814\std{0.006} & --0.283\std{0.009} \\
        \midrule
        LIME-BlackBox                   & \textbf{0.659}\std{0.013} & 0.529\std{0.002} & \textbf{0.131}\std{0.011} & \textbf{0.839}\std{0.013} & 0.589\std{0.002} & 0.250\std{0.013} \\
        Gemini-2.5-Flash                & 0.625\std{0.002} & 0.505\std{0.003} & 0.120\std{0.004} & 0.805\std{0.003} & 0.565\std{0.003} & 0.240\std{0.004} \\
        Gemma-2-9B-It                   & 0.506\std{0.004} & \textbf{0.422}\std{0.008} & 0.083\std{0.005} & 0.714\std{0.001} & \textbf{0.425}\std{0.001} & \textbf{0.289}\std{0.001} \\
        Llama-3.1-8B-It                 & 0.523\std{0.002} & 0.457\std{0.004} & 0.066\std{0.004} & 0.701\std{0.001} & 0.557\std{0.000} & 0.143\std{0.000} \\
        Mistral-7B-It                   & 0.542\std{0.008} & 0.492\std{0.005} & 0.050\std{0.008} & 0.723\std{0.001} & 0.555\std{0.004} & 0.168\std{0.006} \\
        \midrule
        \multicolumn{7}{c}{\textbf{(b) Sentence Cosine Similarity ($\mathcal{F}_{\text{cosine}}$)}} \\
        \midrule
        \rowcolor{gray!20} \textbf{ESCI (Ours)} & 0.486\std{0.002} & 0.463\std{0.002} & 0.023\std{0.004} & 0.586\std{0.002} & 0.563\std{0.002} & 0.023\std{0.003} \\
        \rowcolor{gray!8} StringMatching & 0.597\std{0.004} & \textbf{0.357}\std{0.003} & 0.240\std{0.006} & 0.702\std{0.004} & \textbf{0.455}\std{0.003} & 0.247\std{0.006} \\
        \rowcolor{gray!8} Random (Dir.)  & 0.472\std{0.001} & 0.579\std{0.001} & --0.107\std{0.002} & 0.563\std{0.001} & 0.715\std{0.001} & --0.152\std{0.000} \\
        \midrule
        LIME-BlackBox                   & \textbf{0.685}\std{0.003} & 0.385\std{0.004} & \textbf{0.301}\std{0.007} & \textbf{0.788}\std{0.003} & 0.490\std{0.004} & \textbf{0.298}\std{0.005} \\
        Gemini-2.5-Flash                & 0.491\std{0.002} & 0.458\std{0.002} & 0.033\std{0.003} & 0.592\std{0.002} & 0.555\std{0.002} & 0.037\std{0.003} \\
        Gemma-2-9B-It                   & 0.514\std{0.001} & 0.529\std{0.001} & --0.015\std{0.001} & 0.573\std{0.001} & 0.638\std{0.001} & --0.065\std{0.001} \\
        Llama-3.1-8B-It                 & 0.460\std{0.001} & 0.560\std{0.002} & --0.099\std{0.001} & 0.591\std{0.003} & 0.665\std{0.001} & --0.074\std{0.003} \\
        Mistral-7B-It                   & 0.503\std{0.003} & 0.513\std{0.000} & --0.010\std{0.003} & 0.631\std{0.001} & 0.602\std{0.002} & 0.029\std{0.001} \\
        \midrule
        \multicolumn{7}{c}{\textbf{(c) Normalized Log-Likelihood ($\mathcal{F}_{\text{logprob}}$)}} \\
        \midrule
        \rowcolor{gray!20} \textbf{ESCI (Ours)} & 0.658\std{0.000} & 0.534\std{0.000} & 0.124\std{0.000} & 0.778\std{0.000} & 0.574\std{0.000} & 0.204\std{0.000} \\
        \rowcolor{gray!8} StringMatching & \textbf{0.922}\std{0.000} & \textbf{0.197}\std{0.000} & \textbf{0.725}\std{0.000} & \textbf{0.972}\std{0.000} & \textbf{0.247}\std{0.000} & \textbf{0.725}\std{0.000} \\
        \rowcolor{gray!8} Random (Dir.)  & 0.420\std{0.000} & 0.600\std{0.000} & --0.180\std{0.000} & 0.570\std{0.000} & 0.700\std{0.000} & --0.130\std{0.000} \\
        \midrule
        LIME-BlackBox                   & 0.830\std{0.000} & 0.409\std{0.000} & 0.421\std{0.000} & 0.880\std{0.000} & 0.459\std{0.000} & 0.421\std{0.000} \\
        Gemini-2.5-Flash                & 0.665\std{0.000} & 0.520\std{0.000} & 0.145\std{0.000} & 0.785\std{0.000} & 0.560\std{0.000} & 0.225\std{0.000} \\
        Gemma-2-9B-It                   & 0.471\std{0.000} & 0.527\std{0.000} & --0.056\std{0.000} & 0.621\std{0.000} & 0.607\std{0.000} & 0.014\std{0.000} \\
        Llama-3.1-8B-It                 & 0.436\std{0.000} & 0.583\std{0.000} & --0.147\std{0.000} & 0.586\std{0.000} & 0.663\std{0.000} & --0.077\std{0.000} \\
        Mistral-7B-It                   & 0.510\std{0.000} & 0.574\std{0.000} & --0.064\std{0.000} & 0.660\std{0.000} & 0.654\std{0.000} & 0.006\std{0.000} \\
        \bottomrule
    \end{tabular}%
    }
\end{table}

%% file: Tables/tab_wikibio_faithfulness_whitebox.tex
\begin{table}[h]
    \centering
    \scriptsize
    \caption{\textbf{Faithfulness Evaluation on WikiBio under NLI Entailment (White-Box and Gray-Box).} Comparison against internal-access baselines under the gold-standard metric ($\mathcal{F}_{\text{NLI}}$). Mean and standard deviation across random seeds.}\label{tab:wikibio_faithfulness_whitebox}\label{tab_wikibio_faithfulness_whitebox}
    \resizebox{\columnwidth}{!}{%
    \setlength{\tabcolsep}{2.2pt}
    \renewcommand{\arraystretch}{1.10}
    \begin{tabular}{lc|ccc|ccc}
        \toprule
        & & \multicolumn{3}{c|}{\textbf{Llama-3.1-8B-Instruct}} & \multicolumn{3}{c}{\textbf{Gemma-2-9B-It}} \\
        \cmidrule(lr){3-5} \cmidrule(lr){6-8}
        \textbf{Method} & \textbf{Access} & \textbf{Suff.}$\uparrow$ & \textbf{Comp.}$\downarrow$ & \textbf{Gap}$\uparrow$ & \textbf{Suff.}$\uparrow$ & \textbf{Comp.}$\downarrow$ & \textbf{Gap}$\uparrow$ \\
        \midrule
        \rowcolor{gray!20} \textbf{ESCI (Ours)} & Black-Box & 0.620\std{0.002} & 0.512\std{0.003} & 0.108\std{0.004} & 0.800\std{0.003} & 0.572\std{0.003} & 0.228\std{0.004} \\
        \midrule
        Attention                       & White-Box & 0.725\std{0.013} & 0.632\std{0.005} & 0.093\std{0.018} & \textbf{0.943}\std{0.004} & 0.859\std{0.002} & 0.084\std{0.004} \\
        AttentionRollout                & White-Box & 0.725\std{0.004} & 0.632\std{0.010} & 0.093\std{0.012} & \textbf{0.943}\std{0.007} & 0.859\std{0.003} & 0.084\std{0.010} \\
        GradientSHAP                    & White-Box & 0.609\std{0.014} & 0.499\std{0.005} & 0.111\std{0.019} & 0.817\std{0.003} & 0.656\std{0.004} & 0.160\std{0.003} \\
        Input$\times$Gradient          & White-Box & 0.520\std{0.003} & 0.547\std{0.007} & --0.027\std{0.010} & 0.665\std{0.005} & 0.636\std{0.009} & 0.029\std{0.013} \\
        \midrule
        SentenceLOO                     & Gray-Box  & 0.606\std{0.007} & 0.483\std{0.002} & 0.123\std{0.009} & 0.882\std{0.009} & 0.588\std{0.006} & \textbf{0.295}\std{0.003} \\
        LIME-GrayBox                    & Gray-Box  & \textbf{0.636}\std{0.010} & \textbf{0.487}\std{0.003} & \textbf{0.149}\std{0.007} & 0.816\std{0.010} & \textbf{0.547}\std{0.003} & 0.269\std{0.010} \\
        \bottomrule
    \end{tabular}%
    }
\end{table}

%% file: Tables/tab_hc3_attribution_qualitative.tex
\begin{table}[!t]
    \centering
    \scriptsize
    \caption{\textbf{Comparison of Attribution Scores on HC3.} Highest attributed scores per method are in \textbf{bold}; negligible scores ($\le 0.05$) are \textcolor{gray}{grayed}. The target response sentence is \hlg{highlighted}.}\label{tab:hc3_attribution_qualitative}\label{tab_hc3_attribution_qualitative}
    \resizebox{\columnwidth}{!}{%
    \setlength{\tabcolsep}{3pt}
    \renewcommand{\arraystretch}{1.0}
    \begin{tabular}{@{} c >{\raggedright\arraybackslash}p{5.1cm} c c c >{\raggedright\arraybackslash}p{4.5cm} @{}}
        \toprule
        \textbf{ID} & \textbf{Prompt Sentences} & \textbf{ESCI} & \textbf{GPT-4o} & \textbf{Gemini} & \textbf{Response Context \& Target Sentence} \\
        \midrule
        
        $\mathbf{6}$ & 
        [0] Why say ``half a dozen'' instead of ``six''? \newline
        [1] It seems like such a common occurrence. \newline
        [2] Why take the time and effort to say the extra \dots \newline
        [3] Explain like I'm five.
        & 
        \makecell[t]{\dimsc{0.18} \\ \dimsc{0.00} \\ \dimsc{0.00} \\ \best{0.82}}
        & 
        \makecell[t]{\best{0.70} \\ \dimsc{0.20} \\ \dimsc{0.10} \\ \dimsc{0.00}}
        & 
        \makecell[t]{\best{0.80} \\ \dimsc{0.00} \\ \dimsc{0.00} \\ \dimsc{0.20}}
        & 
        Okay! Imagine you have a box of cookies. \hlg{If you have six cookies, you can just say ``six.''} But if you say ``half a dozen,'' it's like saying ``half of a bigger group'' of cookies. People like to use ``half a dozen'' because it sounds a little fancier\dots
        \\ \midrule

        $\mathbf{21}$ & 
        [0] RGB lines on a picture of your monitor \dots \newline
        [1] Please explain like I'm five.
        & 
        \makecell[t]{\best{0.85} \\ \dimsc{0.15}}
        & 
        \makecell[t]{\best{0.80} \\ \dimsc{0.20}}
        & 
        \makecell[t]{\best{1.00} \\ \dimsc{0.00}}
        & 
        \dots the camera gets a little mixed up and shows the colors in a funny way. \hlg{That's why you see those RGB lines!} \dots
        \\ \midrule

        $\mathbf{33}$ & 
        [0] fuel octane. \newline
        [1] What happens if I feed my Nissan Versa \dots \newline
        [3] Please explain like I'm five.
        & 
        \makecell[t]{\best{0.88} \\ \dimsc{0.12} \\ \dimsc{0.00}}
        & 
        \makecell[t]{\dimsc{0.00} \\ \best{0.70} \\ \dimsc{0.30}}
        & 
        \makecell[t]{\dimsc{0.00} \\ \best{0.70} \\ \dimsc{0.30}}
        & 
        \dots not work anymore! So, it's best to stick with the regular gas (87 octane) that your car is designed to use. \hlg{That way, it will run smoothly and be happy!}
        \\ \midrule

        $\mathbf{48}$ & 
        [0] What those black lines on the road are. \newline
        [1] EDIT: Sorry about the confusion, I meant \dots \newline
        [3] Explain like I'm five.
        & 
        \makecell[t]{\dimsc{0.00} \\ \best{0.76} \\ \dimsc{0.24}}
        & 
        \makecell[t]{\dimsc{0.40} \\ \dimsc{0.10} \\ \best{0.50}}
        & 
        \makecell[t]{\best{0.55} \\ \dimsc{0.00} \\ 0.45}
        & 
        \dots lines show where the lanes are, while others can tell you if you can park or if you need to stop. \hlg{They are like guides that help everyone follow the rules of the road!}
        \\ \midrule

        $\mathbf{94}$ & 
        [0] What's the point of finding planets light years \dots \newline
        [2] Why can't we spend money on improving \dots \newline
        [3] Please explain like I'm five.
        & 
        \makecell[t]{\best{0.72} \\ \dimsc{0.00} \\ \dimsc{0.28}}
        & 
        \makecell[t]{\dimsc{0.00} \\ \best{0.70} \\ \dimsc{0.30}}
        & 
        \makecell[t]{0.30 \\ 0.30 \\ \best{0.40}}
        & 
        \dots while also exploring space, because both are important for our future. \hlg{It's like making sure your toys are clean and also dreaming about getting new ones!}
        \\ \midrule

        $\mathbf{142}$ & 
        [0] The most prominent members of the current \dots \newline
        [1] Explain like I'm five.
        & 
        \makecell[t]{\best{0.78} \\ \dimsc{0.22}}
        & 
        \makecell[t]{\best{0.80} \\ \dimsc{0.20}}
        & 
        \makecell[t]{\best{0.90} \\ \dimsc{0.10}}
        & 
        \dots their own thing and keep the country safe. \hlg{People are talking a lot about these ideas as they get ready to vote!} \dots
        \\ 
        \bottomrule
    \end{tabular}%
    }
\end{table}

%% file: appendix.tex

\section{Energy Network: Pre-training Details}\label{app:ebm_pretraining_details}

\begin{figure}[h]
    \centering
    \includegraphics[width=0.75\columnwidth]{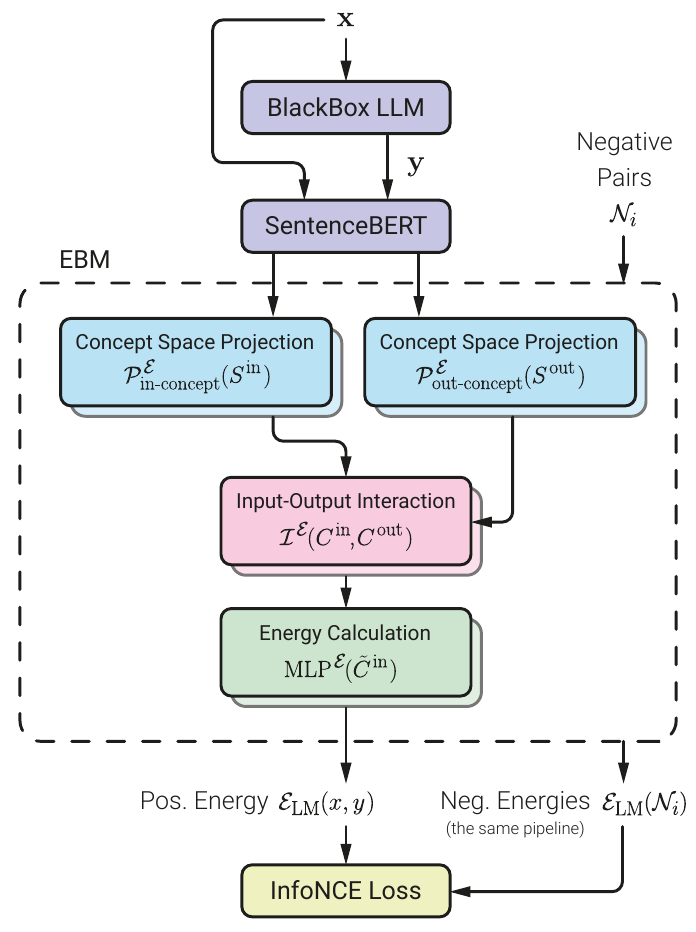}
    \caption{\textbf{Pre-training Pipeline of the EBM Surrogate.} Sentence embeddings are projected into a contextual concept space via self-attention, followed by cross-attention to model prompt-response interactions. An attention-pooling layer and MLP aggregate these representations into a scalar energy $\mathcal{E}_{\text{LM}}(\mathbf{x}, \mathbf{y}; \theta)$. The surrogate is optimized via dual InfoNCE objectives: \textit{fidelity} contrasts authentic pairs against global negatives to anchor the target LLM's generative style, while \textit{local dependency} contrasts partially masked sequences to isolate fine-grained sentence dependencies.}\label{fig:ebm_pretraining_pipeline}
\end{figure}

The pre-training pipeline for the Energy-Based Model is illustrated in Fig.~\ref{fig:ebm_pretraining_pipeline}. All EBM variants were trained on dual NVIDIA T4 GPUs using the AdamW optimizer with a linear learning rate warmup schedule. Each pre-training run required approximately $25$ hours. Table~\ref{tab:ebm_hyperparameters} summarizes the structural specifications and optimization hyperparameters for the surrogate network.

\input{Tables/tab_ebm_hyperparameters}

\vspace{0.25cm}
\paragraph{Model Configurations}
To isolate the empirical contribution of each training objective, we evaluate three distinct EBM training regimes:
(i)~\textbf{Fidelity-only} ($\mathcal{E}_{\text{Fidelity}}$, $\lambda=0$), trained exclusively on global fidelity negatives to assess surrogate capability under macro-level stylistic contrast alone;
(ii)~\textbf{Dependency-only} ($\mathcal{E}_{\text{Dep}}$, $\lambda=1$), trained solely on counterfactual partial dependency contrasts to examine whether localized sentence dependencies can be captured without explicit macro-distributional anchoring; and
(iii)~\textbf{Hybrid} ($\mathcal{E}_{\text{Hybrid}}$, $\lambda=0.9$), our primary proposed framework combining both objectives to balance fine-grained causal dependency learning with macro-distributional regularization.

\vspace{0.25cm}
\paragraph{Negative Sampling Strategies}
The geometry of the learned energy landscape is governed by the composition of the negative candidate sets in the InfoNCE objective (\ref{eq:ebm_total_loss}). For a mini-batch of size $B$ comprising authentic pairs ${(\mathbf{x}_i, \mathbf{y}_i)}_{i=1}^B$, we implement the following sampling protocols:

\vspace{0.15cm}
\noindent\textit{1. Global Fidelity Samplers ($\mathcal{E}_{\text{Fidelity}}$):}
To anchor the surrogate to the target LLM's characteristic generation style, we formulate four negative generation mechanisms:
(a)~\textit{Human Reference} (\texttt{response\_human}): holding prompt $\mathbf{x}_i$ fixed, the LLM response $\mathbf{y}_i$ is replaced with the corresponding human reference $\mathbf{y}^{\text{hum}}_i$ from the benchmark corpus, yielding candidate set $\mathcal{N}^{\text{hum}}_i = \{ (\mathbf{x}_i, \mathbf{y}^{\text{hum}}_i) \}$;
(b)~\textit{Alternative Language Model} (\texttt{response\_other\_lm}): prompt $\mathbf{x}_i$ is paired with a continuation $\mathbf{y}^{\text{other}}_i$ generated by an alternative model (e.g., \texttt{GPT-2-Medium}), giving $\mathcal{N}^{\text{other}}_i = \{ (\mathbf{x}_i, \mathbf{y}^{\text{other}}_i) \}$;
(c)~\textit{Off-Topic Pairings} (\texttt{off\_topic}): one component of the authentic pair is swapped with a randomly selected counterpart from another instance in the batch ($j \ne i$), producing off-topic response sets $\mathcal{N}^{\text{off-topic-resp}}_i = \{ (\mathbf{x}_i, \mathbf{y}_j) \mid j \ne i \}$ and prompt sets $\mathcal{N}^{\text{off-topic-pmt}}_i = \{ (\mathbf{x}_j, \mathbf{y}_i) \mid j \ne i \}$; and
(d)~\textit{Sentence Masking} (\texttt{sentence\_masking}): individual sentences are stochastically ablated with independent dropout probability $p_{\text{drop}} = 0.3$ (conditioned on dropping and retaining at least one sentence each), producing corrupted response candidates $\mathcal{N}^{\text{mask-resp}}_i = \{ (\mathbf{x}_i, \mathbf{y}^{\text{mask}}_i) \}$ and prompt candidates $\mathcal{N}^{\text{mask-pmt}}_i = \{ (\mathbf{x}^{\text{mask}}_i, \mathbf{y}_i) \}$.

\vspace{0.15cm}
\noindent\textit{2. Local Dependency Samplers ($\mathcal{E}_{\text{Dep}}$):}
To evaluate whether specific response segments causally depend on specific prompt segments, all prompts and responses in the batch are first partially masked to produce $(\mathbf{x}_{\text{part}}, \mathbf{y}_{\text{part}})$. We then apply two complementary in-batch counterfactual samplers:

\vspace{0.15cm}
\noindent\textit{Partial Response Dependency} (\texttt{partial\_response\_dep}): Holding partial prompt $\mathbf{x}_{\text{part}, i}$ fixed, the authentic partial response $\mathbf{y}_{\text{part}, i}$ is contrasted against mismatched partial responses from all other batch instances via~\eqref{eq:app_neg_resp_dep}:
\begin{equation}\label{eq:app_neg_resp_dep}
    \resizebox{0.92\columnwidth}{!}{$
    \mathcal{N}^{\text{resp-dep}}_i = \left\{ \left(\mathbf{x}_{\text{part}, i}, \mathbf{y}'_{\text{part}, j}\right) \;\middle|\; j \in \{1, \dots, B\}, j \neq i \right\}
    $}
\end{equation}
This forces the energy surrogate to verify whether the output proposition constitutes a coherent logical continuation of the restricted prompt concepts.

\vspace{0.15cm}
\noindent\textit{Partial Prompt Dependency} (\texttt{partial\_prompt\_dep}): Holding partial response $\mathbf{y}_{\text{part}, i}$ fixed, the authentic partial prompt $\mathbf{x}_{\text{part}, i}$ is contrasted against mismatched partial prompts across the batch via~\eqref{eq:app_neg_pmt_dep}:
\begin{equation}\label{eq:app_neg_pmt_dep}
    \resizebox{0.92\columnwidth}{!}{$
    \mathcal{N}^{\text{pmt-dep}}_i = \left\{ \left(\mathbf{x}'_{\text{part}, j}, \mathbf{y}_{\text{part}, i}\right) \;\middle|\; j \in \{1, \dots, B\}, j \neq i \right\}
    $}
\end{equation}
This ensures that the response is linked directly to its true antecedents rather than generic in-batch topics.

\vspace{0.15cm}
\noindent\textit{3. Composite Hybrid Sampler ($\mathcal{E}_{\text{Hybrid}}$):}
In our primary hybrid architecture, raw \texttt{off\_topic} and \texttt{sentence\_masking} samplers are omitted. Because the counterfactual partial dependency samplers inherently introduce both topical mismatch and structural masking, retaining them separately introduces redundant optimization noise. Thus, the composite negative candidate set for each instance $i$ is formulated as~\eqref{eq:app_neg_hybrid}:
\begin{equation}\label{eq:app_neg_hybrid}
    \resizebox{0.92\columnwidth}{!}{$
    \mathcal{N}^{\text{hybrid}}_i = \mathcal{N}^{\text{resp-dep}}_i \cup \mathcal{N}^{\text{pmt-dep}}_i \cup \mathcal{N}^{\text{hum}}_i \cup \mathcal{N}^{\text{other}}_i
    $}
\end{equation}

\section{Interpreter: Training and Optimization Details}\label{app:interpreter_training_details}

The interpreter $\mathcal{IN}(\mathbf{x}, \mathbf{y}, T; \alpha)$ is trained using the frozen Hybrid ($\lambda=0.9$) EBM as a critic. Table~\ref{tab:interpreter_hyperparameters} outlines the architecture, optimization hyperparameters, and joint training configuration for the interpreter. Across all experiments, end-to-end training of the complete framework (both EBM pre-training and interpreter optimization) required approximately $30$ hours on dual NVIDIA T4 GPUs.

\input{Tables/tab_interpreter_hyperparameters}

\begin{figure}[!t]
    \centering
    \includegraphics[width=\columnwidth]{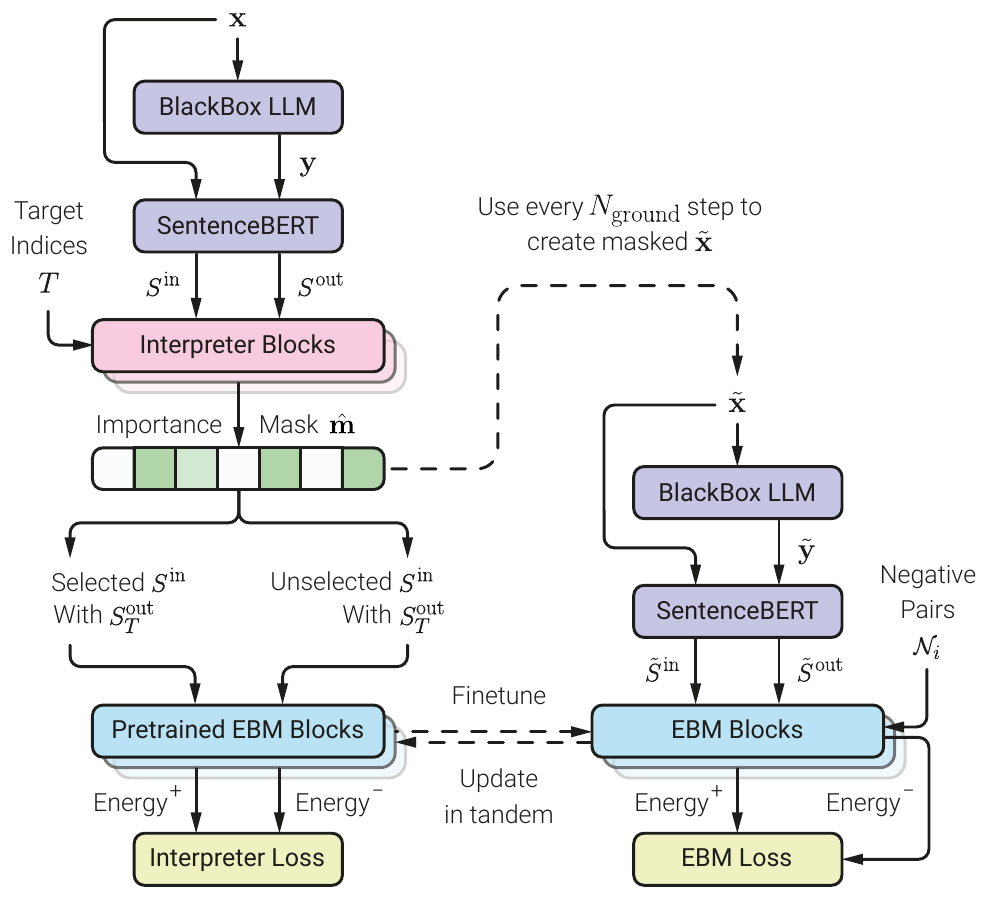}
    \caption{\textbf{Alternating Optimization Protocol.} \textbf{(Left)} Standard phase: The interpreter predicts a soft mask $\hat{\mathbf{m}}$ over prompt sentences; parameters $\alpha$ are updated to maximize the energy gap using the frozen EBM as a critic. \textbf{(Right)} Periodic grounding phase: Every $N_{\text{ground}}$ steps, the black-box LLM is queried with the hard-masked prompt $\mathbf{x}_{\text{hard}}$ to generate an authentic response $\tilde{\mathbf{y}}$, realigning the surrogate EBM with the interpreter's evolving masking policy.}\label{fig:joint_training_pipeline}
\end{figure}

\subsection{Alternating Optimization Protocol}\label{app:interpreter_alternating_optimization}

While Section~\ref{subsec:interpreter} introduces the alternating optimization objective, we provide here the full mathematical formulation of the two alternating update steps. Let $\theta^{(k)}$ and $\alpha^{(k)}$ denote the parameter states of the EBM and interpreter at step $k$, respectively.

\paragraph{Step 1: Interpreter Update}
In standard iterations, the surrogate EBM parameters $\theta^{(k-1)}$ are held frozen. The interpreter generates continuous selection probabilities $\hat{\mathbf{m}} = \text{Sigmoid}(\mathcal{IN}(\mathbf{x}, \mathbf{y}, T; \alpha^{(k-1)})) \in {(0, 1)}^{N_{\text{max}}}$. The prompt is partitioned into the selected subset $\tilde{\mathbf{x}} = \mathbf{x} \odot \hat{\mathbf{m}}$ and the discarded remainder $\mathbf{x} - \tilde{\mathbf{x}} = \mathbf{x} \odot (1 - \hat{\mathbf{m}})$. Parameters $\alpha$ are updated via gradient descent to minimize $\mathcal{L}_{\mathcal{IN}}$ by maximizing the energy gap while penalizing non-sparse selections via~\eqref{eq:app_interpreter_update}:
\begin{equation}\label{eq:app_interpreter_update}
    \resizebox{0.95\columnwidth}{!}{$
    \begin{aligned}
        \alpha^{(k)} \leftarrow \alpha^{(k-1)} - \eta_{\alpha} \nabla_\alpha \Big( & \mathcal{E}_{\text{LM}}(\tilde{\mathbf{x}}, \mathbf{y}_T; \theta^{(k-1)}) \\
        & - \mathcal{E}_{\text{LM}}(\mathbf{x} - \tilde{\mathbf{x}}, \mathbf{y}_T; \theta^{(k-1)}) + \lambda_{\text{sparse}} \|\hat{\mathbf{m}}\|_1 \Big)
    \end{aligned}
    $}
\end{equation}

\paragraph{Step 2: Periodic Grounding and EBM Fine-Tuning}
Every $N_{\text{ground}}$ steps, the system performs periodic grounding to adapt the surrogate to the non-stationary distribution of masked inputs. We apply the deterministic hard mask with threshold $\epsilon$ and query the target black-box LLM API directly via~\eqref{eq:app_hard_mask_query}:
\begin{equation}\label{eq:app_hard_mask_query}
    \mathbf{x}_{\text{hard}} = \mathbf{x} \odot \mathbb{I}(\hat{\mathbf{m}} > \epsilon), \qquad \tilde{\mathbf{y}} = \text{LLM}(\mathbf{x}_{\text{hard}})
\end{equation}
This query yields a synthetic, ground-truth pair $(\mathbf{x}_{\text{hard}}, \tilde{\mathbf{y}})$ reflecting the LLM's generation under the current masking. 

To fine-tune the EBM on this new sample, we minimize the dual objective $\mathcal{L}_{\text{total}}$ (\ref{eq:ebm_total_loss}). Because no human-authored continuation exists for the synthetic prompt $\mathbf{x}_{\text{hard}}$, the fidelity negative set $\mathcal{N}^{\text{fid}}$ is adapted into $\tilde{\mathcal{N}}$ by replacing \texttt{response\_human} with a random base model response to preserve distributional contrast. The EBM parameters are then updated as~\eqref{eq:app_ebm_update}:
\begin{equation}\label{eq:app_ebm_update}
    \resizebox{0.85\columnwidth}{!}{$
    \begin{aligned}
        \theta^{(k)} \leftarrow \theta^{(k-1)} - \eta_{\theta} \nabla_\theta \Big( & (1-\lambda)\mathcal{L}_{\text{fidelity}}(\mathbf{x}_{\text{hard}}, \tilde{\mathbf{y}}, \tilde{\mathcal{N}}) \\
        & + \lambda\mathcal{L}_{\text{dep}}(\mathbf{x}_{\text{hard}}, \tilde{\mathbf{y}}) \Big)
    \end{aligned}
    $}
\end{equation}

\paragraph{Theoretical Rationale}
This periodic grounding protocol fulfills two essential theoretical functions in amortized energy optimization.
First, it \textbf{mitigates adversarial energy basins}: as established by Tu \textit{et al.}~\cite{tu2020improving}, an ungrounded inference network can behave adversarially against a static EBM, exploiting unconstrained, spurious low-energy pockets. Intermittent API queries dynamically eliminate these artificial minima by re-anchoring the surrogate in authentic LLM behavior.
Second, it provides \textbf{dynamic step-unrolling}: following the principles of step-unrolled denoising autoencoders~\cite{savinov2022stepunrolled}, when an upstream selector alters masking patterns throughout training, evaluating against a static noise prior induces severe inference failure. Periodic grounding acts as a step-unrolling mechanism, continually synchronizing the EBM with the interpreter's non-stationary masking distribution.

\input{Tables/tab_hc3_dataset_sample}

\section{Benchmark Datasets and Task Adaptation}\label{app:datasets_adaptation}

To assess post-hoc interpretability across diverse generative contexts, we construct our experimental testbed around two complementary datasets: unconstrained conversational generation (\textbf{HC3}) and constrained factual synthesis (\textbf{WikiBio}).

\vspace{0.25cm}
\paragraph{HC3 (Human ChatGPT Comparison Corpus)}
The HC3 benchmark~\cite{guo2023closechatgpthumanexperts} encompasses tens of thousands of open-domain questions spanning community QA (such as Reddit ELI5), medicine, finance, and law. In this setting, the input prompt is an unconstrained user query, and the target language model generates an explanatory, multi-sentence response. Because open conversational generation lacks deterministic attribution labels, we leverage authentic human-written answers from the dataset as robust contrastive negative samples during surrogate pre-training. This anchors the EBM to the target model's distinct generative style and predictive manifold. Table~\ref{tab:hc3_dataset_sample} displays an illustrative instance from HC3, presenting the user query alongside completions from \texttt{GPT-4o-Mini} and the authentic human baseline.

\input{Tables/tab_wikibio_dataset_sample}

\vspace{0.25cm}
\paragraph{WikiBio}
While HC3 evaluates interpretability in unconstrained conversation, measuring factual fidelity requires an environment where output assertions map directly to explicit input antecedents. We therefore adapt the WikiBio dataset~\cite{lebret2016neural}, which pairs Wikipedia biographical infobox tables with introductory biography articles. 
We linearize each tabular key-value infobox into a series of isolated, atomic natural language statements (e.g., ``The debutyear is 1993.''). The target language model is then prompted with an explicit directive instructing it to synthesize a coherent biographical paragraph relying exclusively on the provided facts, strictly prohibiting conversational pleasantries or parametric knowledge retrieval. Table~\ref{tab:wikibio_dataset_sample} illustrates this pipeline with a representative instance, presenting the prompt, the biography synthesized by \texttt{Llama-3.1-8B-Instruct}, and the reference human passage.

\section{Fuzzy String Matching Quasi-Ground Truth}\label{app:fuzzy_string_matching}

Evaluating explanation faithfulness in natural language generation is fundamentally hindered by the absence of deterministic ground-truth attributions. Because internal autoregressive pathways are opaque, human annotators cannot definitively establish which prompt sentences an LLM relied upon when generating a given output phrase. However, in our constrained WikiBio setting, each input proposition corresponds to an isolated, verifiable factual attribute (e.g., date of birth, team names, or athletic statistics). This structured configuration allows us to formulate an objective lexical quasi-ground truth by mapping generated biographical sentences back to their source propositions. To extract these mappings systematically across thousands of test examples without manual human labeling, we implement an automated two-stage lexical-semantic matching pipeline powered by the \texttt{spaCy} library~\cite{honnibal2017spacy}.

\vspace{0.25cm}
\paragraph{Linguistic Preprocessing}
Given an input prompt $\mathbf{x} = \{s_1, s_2, \dots, s_N\}$ composed of $N$ factual propositions and a generated response $\mathbf{y} = \{y_1, y_2, \dots, y_M\}$ consisting of $M$ output sentences, both sequences are processed through \texttt{spaCy}~\cite{honnibal2017spacy} for sentence boundary detection, subword tokenization, and part-of-speech (POS) tagging. Punctuation marks and standard closed-class stop words are filtered out, and lemmatization is applied to extract core lexical roots. Informational content extraction specifically targets proper nouns (\texttt{PROPN}), common nouns (\texttt{NOUN}), numerical digits and calendar dates (\texttt{NUM}), main verbs (\texttt{VERB}), and descriptive adjectives (\texttt{ADJ}).

\vspace{0.25cm}
\paragraph{Two-Stage Matching Pipeline}
For each target generated sentence $y_j$, the algorithm identifies corresponding causal prompt antecedents $s_i$ via a hierarchical two-stage strategy:

\vspace{0.15cm}
\noindent\textit{Stage 1 (Exact Content-Entity Intersection):} The algorithm computes the exact set intersection of content lemmas and named entity spans between prompt proposition $s_i$ and generated sentence $y_j$. If distinctive entities (such as proper names, sports teams, numerical values, or calendar dates) appear identically in both sentences, a deterministic antecedent match is recorded.

\vspace{0.15cm}
\noindent\textit{Stage 2 (Fuzzy String Matching):} Strict string intersection frequently fails when language models introduce morphological adaptations, syntactic rearrangements, or minor parametric hallucinations (e.g., generating ``1953'' instead of ``1954'', altering singular/plural inflections, or rephrasing team descriptors). To capture these soft alignments, we apply character-level fuzzy string matching using the Levenshtein similarity ratio across candidate token pairs. Any proposition--sentence pair achieving a normalized fuzzy similarity score $\tau \ge 0.75$ is resolved as an active attribution link.

\vspace{0.25cm}
\paragraph{Heuristic Score Assignment}
For each target sentence $y_j$, the attribution score assigned to prompt sentence $s_i$ is set to $1$ if $s_i$ satisfies either exact entity intersection or above-threshold fuzzy matching, and $0$ otherwise (subsequently normalized across active antecedents such that $\sum_{i=1}^N a_{i,j} = 1.0$). 
While this \texttt{StringMatching} procedure serves as an objective and transparent reference baseline for lexical alignment, it operates purely at the surface level. As demonstrated in our experiments (Section~\ref{sec:experiments}), surface lexical alignment does not always correspond to an LLM's true internal causal reliance, underscoring the necessity of surrogate energy modeling.

\section{Sentence-Level Adaptation of LIME}\label{app:lime_sentence_adaptation}

Local Interpretable Model-agnostic Explanations (\texttt{LIME})~\cite{ribeiro2016lime} is a canonical post-hoc interpretability framework designed to approximate the local decision boundary of a complex predictive model $f$ using an interpretable surrogate $g \in \mathcal{G}$ (typically a sparse linear regressor). However, standard LIME implementations operate at the individual subword token level. In open-ended natural language generation, token-level perturbations induce severe syntactic corruption, high distribution shift, and noisy statistical artifacts that fail to reflect coherent propositional reasoning.

To establish an equitable and mathematically coherent baseline against ESCI, we elevate LIME's perturbation and surrogate modeling to sentence granularity. This ensures that features represent integrated conceptual propositions, enabling a direct scientific comparison between local surrogate regression (\texttt{LIME}) and global energy landscape attribution (ESCI). Below, we detail the mathematical formulation at sentence granularity (Appendix~\ref{app:lime_formulation}) and across both gray-box (Appendix~\ref{app:lime_graybox}) and black-box (Appendix~\ref{app:lime_blackbox}) access regimes.

\subsection{Mathematical Formulation at Sentence Granularity}\label{app:lime_formulation}

Let input prompt $\mathbf{x}$ be partitioned into $N$ distinct propositional sentences according to~\eqref{eq:lime_partition}:
\begin{equation}\label{eq:lime_partition}
    X = \{s_1, s_2, \dots, s_N\}
\end{equation}
In our adapted formulation, each feature corresponds to an entire sentence $s_i$. To measure feature importance, \texttt{LIME} generates $K$ perturbed input instances by sampling binary mask vectors $\mathbf{m}^{(k)} \in {\{0, 1\}}^N$ for $k \in \{1, \dots, K\}$, where $\mathbf{m}^{(k)}_i = 1$ indicates that sentence $s_i$ is preserved in the prompt, and $\mathbf{m}^{(k)}_i = 0$ indicates that it is physically ablated.

Although sentence-level representations reduce feature dimensionality ($N \approx 15$--$24$) compared to long token sequences, the combinatorial space of perturbations $\mathcal{O}(2^N)$ remains large. To guarantee statistical degrees of freedom for the regression while bounding inference compute, we sample the perturbation budget according to~\eqref{eq:lime_sample_budget}:
\begin{equation}\label{eq:lime_sample_budget}
    K = \min\left(2^N, 1000\right)
\end{equation}
For each binary mask $\mathbf{m}^{(k)}$, the corresponding perturbed prompt text $\tilde{X}^{(k)}$ is constructed by concatenating the surviving sentences according to~\eqref{eq:lime_perturbed_text}:
\begin{equation}\label{eq:lime_perturbed_text}
    \tilde{X}^{(k)} = \operatorname{Concat}\left(\{s_i \mid \mathbf{m}^{(k)}_i = 1, \; \forall i\}\right)
\end{equation}

\paragraph{Kernel Weighting}
Following \texttt{LIME}'s local surrogate theory, the linear surrogate model must give greater weight to perturbations that lie close to the original unperturbed input. Distance $d_k$ is computed directly in the binary mask space relative to the unperturbed baseline mask $\mathbf{1} \in {\{0, 1\}}^N$ using cosine distance. Because this metric is normalized in $[0, 1]$, sample weights $w_k$ are assigned using an exponential kernel with fixed bandwidth $\sigma = 0.25$ via~\eqref{eq:lime_kernel_weight}:
\begin{equation}\label{eq:lime_kernel_weight}
    w_k = \exp\left(-\frac{d_k^2}{\sigma^2}\right), \quad \text{with } \sigma = 0.25
\end{equation}

\paragraph{Surrogate Optimization via LassoCV}
Given a scalar dependent target $y_k$ capturing the target model's response to perturbed prompt $\tilde{X}^{(k)}$, an interpretable linear model $g(\mathbf{m}) = \mathbf{m} \cdot \beta$ is optimized by minimizing the weighted Lasso objective~\eqref{eq:lime_lasso_objective}:
\begin{equation}\label{eq:lime_lasso_objective}
    \min_{\beta} \sum_{k=1}^K w_k {\left( y_k - \mathbf{m}^{(k)} \cdot \beta \right)}^2 + \lambda \|\beta\|_1
\end{equation}
where $\beta \in \mathbb{R}^N$ represents the learned attribution coefficients (importance scores for each prompt sentence), and $\lambda$ is an $L_1$ regularization penalty enforcing sparsity. To prevent overfitting and avoid arbitrary manual tuning of $\lambda$, we perform 5-fold cross-validation (\texttt{LassoCV}) across the perturbed sample set to determine the optimal regularization penalty.

The critical methodological distinction between gray-box and black-box LIME lies in how the scalar dependent variable $y_k$ is extracted.

\subsection{Gray-Box LIME (\texorpdfstring{\upshape\texttt{LIME-GrayBox}}{LIME-GrayBox})}\label{app:lime_graybox}

In open-weights models (such as \texttt{Llama-3.1-8B-Instruct} and \texttt{Gemma-2-9B-It}), internal output logits are accessible. In this gray-box setting, we evaluate the impact of prompt sentence ablation on the target LLM's autoregressive log-likelihood of generating the target output sentence $y_T = (v_1, v_2, \dots, v_L)$ comprising $L$ tokens.

To compute this efficiently without expensive token-by-token generation for every perturbation, we employ teacher forcing in a single forward pass. The perturbed prompt tokens and target sentence tokens are concatenated ($\mathbf{I}^{(k)} = [\tilde{X}^{(k)} \oplus y_T]$) and fed into the model. The dependent variable $y_k$ is then computed as the total log-probability of $y_T$, equivalent to the negative cumulative cross-entropy (CE) loss across the target token sequence via~\eqref{eq:lime_graybox_target}:
\begin{equation}\label{eq:lime_graybox_target}
\resizebox{0.95\columnwidth}{!}{$
\begin{aligned}
    y_k &= \log P\left(y_T \mid \tilde{X}^{(k)}\right) \\
    &= \sum_{j=1}^L \log P\left(v_j \mid v_{<j}, \tilde{X}^{(k)}\right) = -\sum_{j=1}^L \text{CE}\left(z_{t+j-1}, v_j\right)
\end{aligned}
$}
\end{equation}
where $t$ is the token length of the perturbed prompt $\tilde{X}^{(k)}$, and $z$ denotes the output logit vectors.

While this formulation perturbs input features at sentence granularity, the dependent variable remains tied to exact token probabilities. Consequently, while \texttt{LIME-GrayBox} provides an accurate measurement of autoregressive token likelihood sensitivity, it is inherently vulnerable to paraphrasing: if an ablated prompt causes the model to express the same underlying concept using alternative vocabulary, token log-likelihood drops sharply, causing the method to misinterpret paraphrastic invariance as an attribution failure. Nevertheless, it represents a strong computational upper bound for local perturbation methods under internal model access.

\subsection{Black-Box LIME (\texorpdfstring{\upshape\texttt{LIME-BlackBox}}{LIME-BlackBox})}\label{app:lime_blackbox}

In proprietary, API-mediated deployments (such as \texttt{GPT-4o-Mini}), internal logits and token distributions are strictly inaccessible. In this black-box setting, each perturbed prompt $\tilde{X}^{(k)}$ is sent directly to the model API to generate a fresh completion $\tilde{y}^{(k)}$. We must then evaluate how effectively $\tilde{y}^{(k)}$ preserves the propositional content of the original target sentence $y_T$.

A tempting heuristic would be computing cosine similarity between dense sentence embeddings of $y_T$ and $\tilde{y}^{(k)}$ (e.g., using \texttt{SentenceTransformers}). However, we strictly avoid this formulation for two theoretical and methodological reasons.
First, \textbf{incompatibility with local additive surrogates}: LIME is theoretically premised on identifying localized, feature-level contributions. Dense sentence embeddings compress an entire passage into a monolithic latent vector, abstracting away localized lexical and syntactic cues and distorting the linear surrogate's additive assumptions.
Second, \textbf{circular evaluation bias (data leakage)}: our downstream empirical benchmarks evaluate explanation faithfulness using semantic similarity metrics (specifically embedding cosine similarity and NLI entailment). If LIME were trained to optimize cosine similarity during baseline score extraction, it would obtain an unfair, circular advantage on downstream evaluations, invalidating comparative analysis.

To circumvent both pitfalls while maintaining local semantic precision, we formulate the dependent variable using \textbf{\texttt{BERTScore-F1}}~\cite{zhang2020bertscore} as defined in~\eqref{eq:lime_blackbox_target}:
\begin{equation}\label{eq:lime_blackbox_target}
    y_k = \text{BERTScore-F1}\left( \tilde{y}^{(k)}, y_T \right)
\end{equation}
\texttt{BERTScore} computes pairwise cosine similarities across contextual token embeddings extracted from a pre-trained transformer (\texttt{RoBERTa}) and performs greedy bipartite matching. This allows \texttt{LIME-BlackBox} to account for paraphrasing and semantic equivalence while preserving token-level sensitivity and preventing circular metric contamination.

%% file: Tables/tab_ebm_hyperparameters.tex
\begin{table}[!t]
    \centering
    \scriptsize
    \renewcommand{\arraystretch}{0.95}
    \caption{\textbf{Architectural specifications and hyperparameters for the surrogate EBM ($\mathcal{E}_{\text{LM}}$).}}\label{tab:ebm_hyperparameters}\label{tab_ebm_hyperparameters}
    \begin{tabular}{lc}
        \toprule
        \textbf{Parameter} & \textbf{Value} \\
        \midrule
        \multicolumn{2}{c}{\textit{Architecture}} \\
        Base Sentence Encoder & all-mpnet-base-v2 \\
        Padding Strategy & Learnable placeholder embeddings \\
        Max Sentence Count ($N_{\text{max}}$) & $20$ (HC3) / $24$ (WikiBio) \\
        Latent Dimension ($d_{\text{model}}$) & $768$ \\
        Self-Attention Layers & $2$ \\
        Cross-Attention Layers & $8$ \\
        Attention Heads & $8$ \\
        Activation Function & GELU \\
        Dropout Rate & $0.1$ \\
        MLP Layers (Energy Head) & $2$ \\
        MLP Hidden Expansion Factor & $2$ \\
        \midrule
        \multicolumn{2}{c}{\textit{Optimization \& Loss}} \\
        Batch Size ($B$) & $16$ \\
        Epochs & $50$ \\
        Learning Rate & $3 \times 10^{-5}$ \\
        Learning Rate Scheduler & Linear with $200$ warmup steps \\
        Local Dependency Weight ($\lambda$) & $0.9$ \\
        InfoNCE Temperature ($\tau$) & $0.1$ \\
        Energy Margin & $0.5$ \\
        \midrule
        \multicolumn{2}{c}{\textit{Dataset}} \\
        Dataset Size & $20,000$ samples \\
        Validation Split & $10\%$ \\
        \bottomrule
    \end{tabular}
\end{table}

%% file: Tables/tab_interpreter_hyperparameters.tex
\begin{table}[!t]
    \centering
    \scriptsize
    \renewcommand{\arraystretch}{0.95}
    \caption{\textbf{Architectural specifications and hyperparameters for the interpreter network ($\mathcal{IN}$).}}\label{tab:interpreter_hyperparameters}\label{tab_interpreter_hyperparameters}
    \begin{tabular}{lc}
        \toprule
        \textbf{Parameter} & \textbf{Value} \\
        \midrule
        \multicolumn{2}{c}{\textit{Architecture \& Selection}} \\
        Concept Space Projection & Initialized from EBM, trainable \\
        Self-Attention Layers & $2$ \\
        Cross-Attention Layers & $8$ \\
        Attention Heads & $8$ \\
        MLP Hidden Dimension & $256$ \\
        Score Normalization & Sigmoid (independent Bernoulli) \\
        Soft Mask Application & Element-wise ($\tilde{\mathbf{x}} = \mathbf{x} \odot \hat{\mathbf{m}}$) \\
        Sparsity Regularization & $L_1$ penalty ($\lambda_{\text{sparse}} = 0.005$) \\
        \midrule
        \multicolumn{2}{c}{\textit{Optimization}} \\
        Epochs & $15$ \\
        Batch Size ($B$) & $16$ \\
        Learning Rate ($\eta_\alpha$) & $1 \times 10^{-5}$ \\
        Energy Margin & $1.0$ \\
        InfoNCE Temperature ($\tau$) & $0.1$ \\
        \midrule
        \multicolumn{2}{c}{\textit{Joint Training \& Grounding}} \\
        EBM Fine-Tuning & Enabled \\
        Grounding Schedule & Once every $5$ epochs \\
        Fine-Tuning Steps per Query & $5$ steps \\
        EBM Fine-Tuning LR ($\eta_\theta$) & $5 \times 10^{-6}$ \\
        Fine-Tuning Energy Margin & $0.5$ \\
        Max Sentence Count ($N_{\text{max}}$) & $20$ (HC3) / $24$ (WikiBio) \\
        \bottomrule
    \end{tabular}
\end{table}

%% file: Tables/tab_hc3_dataset_sample.tex
\begin{table*}[t]
    \centering
    \small
    \caption{\textbf{Representative Sample from the HC3 Dataset.} An open-domain question from the ELI5 subset alongside the generated completion from \texttt{GPT-4o-Mini} and the corresponding human reference answer.}\label{tab:hc3_dataset_sample}\label{tab_hc3_dataset_sample}
    \renewcommand{\arraystretch}{1.2}
    \begin{tabularx}{\textwidth}{@{} >{\bfseries}p{3.2cm} X @{}}
        \toprule
        Component & Text \\
        \midrule
        Question & 
        Why we experience the sensation of falling in our dreams / when we wake up. Title basically says it all. What causes this sensation? Why is it so seemingly universal and why is it so vivid relative to other physical sensations we experience in our dreams? Please explain like I'm five. \\ 
        \midrule
        GPT-4o-Mini Answer & 
        Okay! Imagine your brain is like a big movie theater that shows dreams while you sleep. Sometimes, when you're dreaming, your body might feel like it's falling. This can happen because when you're asleep, your brain is still working hard, and it can get mixed up about what's real and what's not. \par\vspace{0.15cm}
        When you dream about falling, it's like your brain is playing a trick on you! It might be because your body is relaxing and your muscles are letting go, and your brain thinks, ``Whoa, I'm falling!'' This feeling can be really strong and surprising, just like a roller coaster ride! \par\vspace{0.15cm}
        A lot of people feel this way in their dreams because everyone's brain works in similar ways when they sleep. It's like a common movie that lots of people watch in their sleep! \\
        \midrule
        Human Answer & 
        When you're in between deep sleep and the lighter sleep where it's easy to wake up, the sleep paralysis that stops you from acting out your dreams can do strange shit. Sometimes you'll be surfacing to wakefulness again when your brain accidently throws the sleep paralysis switch again. The sudden tension in the body can feel like falling to a sleep-frazzled mind, and it galvanizes you strongly it can even create the sensation you'd dropped onto the bed and bounced. You didn't. Your body just got switched into safe mode by accident. \\
        \bottomrule
    \end{tabularx}
\end{table*}

%% file: Tables/tab_wikibio_dataset_sample.tex
\begin{table*}[ht]
    \centering
    \small
    \caption{\textbf{Representative Sample from the Adapted WikiBio Dataset.} Tabular infobox attributes transformed into isolated natural language propositions, followed by the synthesized biography from \texttt{Llama-3.1-8B-Instruct} and the authentic human reference.}\label{tab:wikibio_dataset_sample}\label{tab_wikibio_dataset_sample}
    \renewcommand{\arraystretch}{1.2}
    \begin{tabularx}{\textwidth}{@{} >{\bfseries}p{3.2cm} X @{}}
        \toprule
        Component & Text \\
        \midrule
        Synthesized Prompt & 
        Without including any conversational preamble, external knowledge, or data corrections, immediately synthesize a biographical paragraph about Craig Paquette using exclusively the provided information. \par\vspace{0.15cm}
        \textbf{Information:} \par
        The finalyear is 2003. \par
        The bats is right. \par
        The birth place is long beach, california. \par
        The debutyear is 1993. \par
        The finaldate is april 24. \par
        The statvalue is .239 99 377. \par
        The position is third baseman / outfielder. \par
        The finalteam is kansas city royals. \par
        The statlabel is batting average home runs runs batted in. \par
        The debutdate is june 1. \par
        The birth date is 28 march 1969. \par
        The article title is craig paquette. \par
        The debutteam is oakland athletics. \par
        The name is craig paquette. \par
        The throws is right. \par\vspace{0.15cm}
        \textbf{Biography:} \\
        \midrule
        Llama-3.1-8B-Instruct Answer & 
        Craig Paquette was born on March 28, 1968. He made his professional baseball debut with the Oakland Athletics on June 1, 1992 as a third baseman/outfielder and threw right-handed. Over the course of his career, which spanned from 1994 to 2005, he played for several teams before joining the Kansas City Royals, where he concluded his playing days on April 24, 2006. During his time at bat, Paquette achieved notable statistics, boasting a batting average of .249, hitting 379 home runs, and accumulating 999 runs bated-in. \\
        \midrule
        Human Answer & 
        Craig Harold Paquette (born march 28, 1969), is a former professional baseball player who played third base in the major leagues from 1993--2003. He last played in for the camden riversharks. He attended rancho alamitos high school and then went on to play college baseball for golden west college. \\
        \bottomrule
    \end{tabularx}
\end{table*}